\documentclass[letterpaper]{article} 
\usepackage{aaai23}  
\usepackage{times}  
\usepackage{helvet}  
\usepackage{courier}  
\usepackage[hyphens]{url}  
\usepackage{graphicx} 
\usepackage{natbib}  
\usepackage{caption} 
\usepackage{algorithm}
\usepackage[noend]{algorithmic}

\usepackage{newfloat}
\usepackage{listings}
\DeclareCaptionStyle{ruled}{labelfont=normalfont,labelsep=colon,strut=off} 
\floatstyle{ruled}
\newfloat{listing}{tb}{lst}{}
\floatname{listing}{Listing}
\usepackage{amsmath}
\usepackage{amssymb}
\usepackage{mathtools}
\usepackage{amsthm}
\usepackage{tabularx}
\usepackage{xltabular}
\usepackage[usenames,dvipsnames]{xcolor}
\usepackage{microtype}
\usepackage{graphicx}
\usepackage{subfigure}
\usepackage{multirow}
\usepackage{siunitx}
\usepackage{booktabs} 
\usepackage{longtable}
\theoremstyle{plain}

\theoremstyle{definition}

\theoremstyle{remark}

\usepackage{fancyhdr}
\definecolor{citecolor}{HTML}{0071bc}
\title{Low Emission Building Control with\\ Zero-Shot Reinforcement Learning}
\author {
    Scott Jeen,\textsuperscript{\rm 1 \rm 3}
    Alessandro Abate, \textsuperscript{\rm 2 \rm 3}
    Jonathan M. Cullen \textsuperscript{\rm 1}
}
\affiliations {
    \textsuperscript{\rm 1} University of Cambridge\\
    \textsuperscript{\rm 2} University of Oxford\\
    \textsuperscript{\rm 3} Alan Turing Institute\\
    \{srj38, jmc99\}@cam.ac.uk \& alessandro.abate@cs.ox.ac.uk
}
\usepackage{bibentry}

\begin{document}

\maketitle

\begin{abstract}
Heating and cooling systems in buildings account for 31\% of global energy use, much of which are regulated by Rule Based Controllers (RBCs) that neither maximise energy efficiency nor minimise emissions by interacting optimally with the grid. Control via Reinforcement Learning (RL) has been shown to significantly improve building energy efficiency, but existing solutions require access to building-specific simulators or data that cannot be expected for every building in the world. In response, we show it is possible to obtain emission-reducing policies without such knowledge \textit{a priori} -- a paradigm we call zero-shot building control. We combine ideas from system identification and model-based RL to create PEARL (\textbf{P}robabilistic \textbf{E}mission-\textbf{A}bating \textbf{R}einforcement \textbf{L}earning) and show that a short period of active exploration is all that is required to build a performant model. In experiments across three varied building energy simulations, we show PEARL outperforms an existing RBC once, and popular RL baselines in all cases, reducing building emissions by as much as 31\% whilst maintaining thermal comfort. Our source code is available online via {\fontfamily{qcr}\selectfont https://enjeeneer.io/projects/pearl/}.
\end{abstract}

\section{Introduction} \label{section: intro}
Heating and cooling systems in buildings account for 31\% of global energy use, primarily in managing occupant thermal comfort and hygiene \cite{cullen2010}. Such systems are usually regulated by rule-based controllers (RBCs) that take system temperature as input, use a temperature setpoint as an objective, and actuate equipment to minimise the error between objective and current state. Whilst usefully simple, RBCs do not maximise energy efficiency, nor can they perform \textit{demand response}: the manipulation of power consumption to better match demand with supply. New techniques that demonstrate this capability across generalised settings would prove valuable climate change mitigation tools.

An option for obtaining control polices across complex, unknown settings is Reinforcement Learning (RL) \cite{Sutton2018}. Where existing advanced control techniques, like the receding-horizon architecture Model Predictive Control (MPC), require the specification of a dynamical model that can be expensive to obtain \cite{morari2017}, RL's strength is in obtaining polices \textit{tabula rasa}, and updating their parameters \textit{online} as the environment evolves. Recent successes in complex physics tasks \cite{lillicrap2015}, gaming \cite{silver2017}, and robotics \cite{gu2017} have highlighted these characteristics. With this generality, one can imagine RL agents being placed in \textit{any} energy-intensive setting, and feasibly learning to control them more efficiently, minimising emissions.

Recent applications of RL to building control have indeed shown marked energy efficiency improvements over conventional controllers. \citet{CHEN2018195} used Q-learning to control HVAC and window actuation in a residential building with 23\% less energy than the existing RBC. Similarly, \citet{zhang2019whole} used Asynchornous Advantage Actor Critic (A3C) to reduce heating demand in a real office building by 16.7\%. However, these contributions are limited by their deployment paradigm, which is:
\begin{enumerate}
    \item \textbf{Simulate} the real-world environment's transition dynamics $p(s_{t+1} | s_t, a_t)$ using a physics-based/generative model; 
    \item \textbf{Pre-train} the agent by sampling the simulator sequentially and updating parameters until convergence; and
    \item \textbf{Deploy} the pre-trained agent in the real-world environment.
\end{enumerate}
We contend that the need to \textit{simulate} the environment \textit{a priori} limits real-world scalability. In the context of building control, creating an accurate simulator in \textit{EnergyPlus} \cite{energyplus2001}, the favoured software, can take an expert months, and is impossible without knowledge of the building topology and thermal parameters. An alternative, is to deploy agents in the real environment \textit{without} pre-training, granting them a short commissioning period of 3 hours to collect data and model the state-action space \cite{LAZIC2018}. We call this \textbf{zero-shot} building control after \citet{socher2013}'s use for out-of-distribution image classification. Zero-shot control could scale to any building given sufficient sensor installation, but \citet{LAZIC2018}'s agent only improved cooling costs by 9\%, poorer performance than the agents trained in simulation. To maximise the emission-abating potential of RL building control, we need new systems that can elicit the performance of pre-trained agents whilst being deployed zero-shot.

In this paper, our primary contribution is showing that deep reinforcement learning algorithms can find performant building control policies \textit{online}, without pre-training. We achieve this with a new approach called PEARL, showing it can reduce annual emissions by up to 31.46\% compared with an RBC whilst maintaining thermal comfort. PEARL is simple to commission, requiring no historical data or simulator access \textit{a priori}, and capable of generalising across varied building archetypes. The scaled deployment of such systems could prove a cost-effective method for tackling climate change.

\section{Related Work} \label{section: related work}
Previous research on RL for building control has focused mainly on model-free algorithms \cite{yu2021}. In this setting, the agent uses data collected from the environment to learn a policy $\pi$ that maps states $s_t$ from a state-space $\mathcal{S}$ to a probability distribution over the action-space $\mathcal{A}$ i.e. $\pi: \mathcal{S} \rightarrow \mathcal{P}(\mathcal{A})$. To obtain an optimal policy, the agent must necessarily visit many states $s \in \mathcal{S}$ and trial many actions $a \in \mathcal{A}$. Doing so is data inefficient, with Deep-Q Learning \cite{mnih2013}, Deep Deterministic Policy Gradient (DDPG) \cite{lillicrap2015}, and Proximal Policy Optimisation (PPO) \cite{schulman2017} each taking in the order 10$^7$ samples in complex, simulated environments to obtain optimal polices. Such data inefficiency has been corroborated in the building control literature. \citet{Wei2017} train a Deep-Q agent to control the HVAC equipment of a 5-zone building with 35\% reduction in energy cost, but pre-train for 8 years in a simulator. Similarly, \citet{VALLADARES2019} require 10 years of simulated data to pre-train a Double-Q agent to control the HVAC in a university classroom with 5\% energy savings. Such data-intensive pre-training can only be achieved in bespoke building simulators that are time-consuming to create, or impossible to specify, in most cases.

In contrast, model-based RL algorithms have demonstrated better sample efficiency. Here, the agent learns the system dynamics mapping from current state-action pair $(s_{t}, a_t)$ to next state $s_{t+1}$, often called the agent's \textit{model} i.e. $f_\theta: (s_t, a_t) \rightarrow s_{t+1}$. Between interactions with the environment, the agent samples its model to create additional data for updating the policy, or to predict the expected reward of a range of candidate action sequences $a_{t:t+H-1}$ to time horizon $H$. Either procedure reduces the agent's reliance on samples from the environment, improving data efficiency.

A popular choice of function approximator are Gaussian Processes (GP), which offer uncertainty quantification and work well with small datasets \cite{williams2006}. PILCO uses GPs, showing state-of-the-art data efficiency on robotic tasks \cite{deisenroth2011pilco}, and \citet{jain2018} used a similar approach to curtail hotel energy-use during a simulated demand response event. However, inference using a data set of size $n$ has complexity $\mathcal{O}(n^3)$ which becomes intractable with more than a few thousand samples \cite{hensman2013}, limiting their applicability for modelling building transition functions with arbitrarily large training sets.

An alternative is to model the transition function using Deep Neural Networks (DNNs). \citet{nagabandi2018} combined DNNs with MPC to solve the \textit{Swimmer} task on the MuJoCo benchmark using 20x fewer datapoints than a model-free approach. In the building control setting, \citet{zhang2019} built a similar agent that reduced energy consumption in a data centre by 21.8\% with 10x fewer training steps than a model-free algorithm. \citet{jain20} then tested the algorithm in-situ, finding an 8\% energy reduction. Despite encouraging data efficiency, the performance of such agents is hampered by overfitting, or \textit{model bias}, in the low-data regime, causing poor generalisation to unobserved transitions. \citet{ding2020} attempt to mitigate model bias by deploying an ensemble of DNNs to model the transition function of a large multi-zone building showing they can achieve 8.2\% energy savings in 10.5x fewer timesteps than model-free approaches. Although the ensemble allows for the quantification of \textit{epistemic} uncertainty, \textit{aleatoric} uncertainty is not captured, potentially limiting performance.

To the best of our knowledge, \citet{LAZIC2018} is the only work that attempts to learn a zero-shot building control policy by interacting with the real environment. Their agent fits a linear model of a datacentre's thermal dynamics using data obtained in a three-hour commissioning period, and selects actions by optimising planned trajectories using MPC. During commissioning, the agent explores the state-action space by performing a uniform random walk in each control variable, bounded to a safe operating range informed by historical data. Their choice of model expedites learning, but limits agent performance as the building's non-linear dynamics are erroneously linearised. In this study we aim to preserve the data efficiency of this approach whilst improving performance with expressive deep networks.

\begin{figure*}[ht!]
    \includegraphics[width=\textwidth]{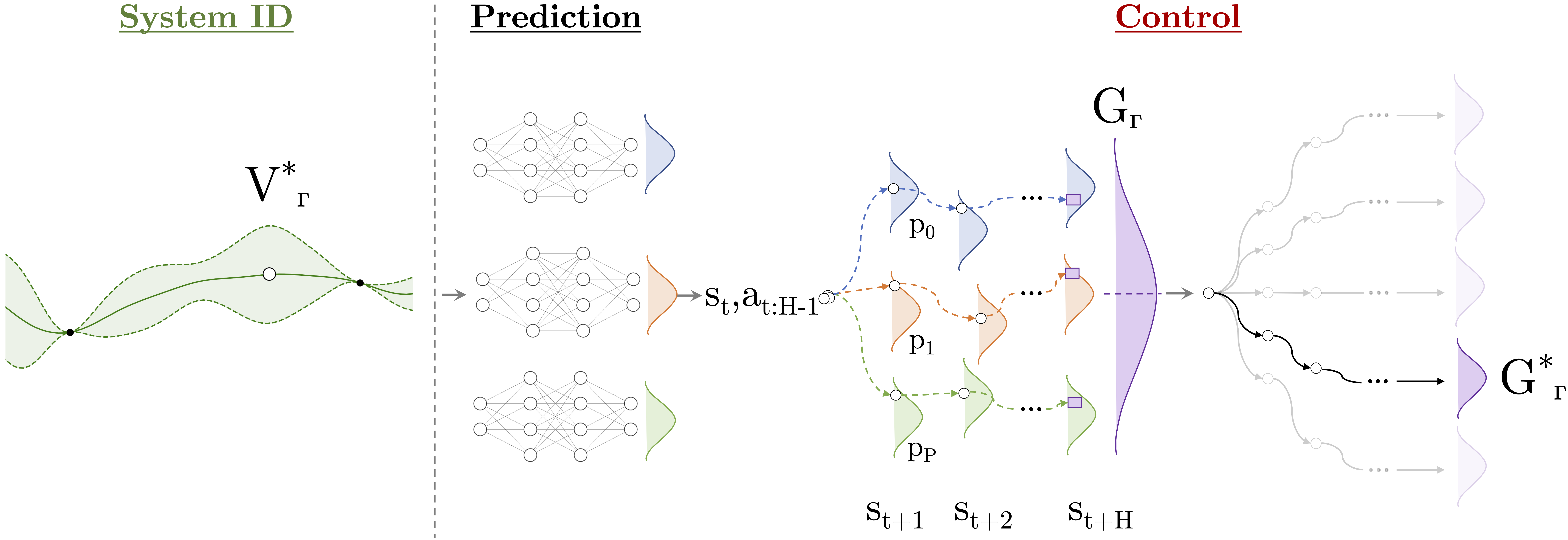}
    \caption{\textbf{PEARL:} our end-to-end deep reinforcement learning approach. \textbf{{\color{OliveGreen}System ID:}} the agent takes actions to explore parts of the state-space with highest predictive variance $\circ = V_\Gamma^*$ to maximise information gain. \textbf{Prediction:} system dynamics modelled as an ensemble of probabilistic deep neural networks. \textbf{{\color{BrickRed}Control:}} trajectory sampling used to predict future rewards $G_\Gamma$ of one action sequence $a_{t:H-1}$, which is compared with many others to find the trajectory with optimal return $G^*_\Gamma$.}
    \label{fig:pearl}
\end{figure*}

\section{PEARL: Probabilistic Emission-Abating Reinforcement Learning} \label{section: PEARL}
\textbf{Problem formulation.} We consider the standard reinforcement learning setup in which an agent takes actions in an environment at discrete timesteps to maximise the cumulative sum of future rewards. We model the environment as an infinite-horizon Markov Decision Process (MDP) characterised by a tuple $(\mathcal{S}, \mathcal{A},\mathcal{R}, \mathcal{T}, \gamma$), where $\mathcal{S} \in \mathbb{R}^{d_s}$ and $\mathcal{A} \in \mathbb{R}^{d_a}$ are continuous state and action spaces, $\mathcal{R}: \mathcal{S} \times \mathcal{A} \rightarrow \mathbb{R}$ is a reward function,  $\mathcal{T}: \mathcal{S} \times \mathcal{A} \times \mathcal{S} \rightarrow \mathbb{R}$ is a transition function, and $\gamma \in [0,1]$ is a discount parameter. In our case we do not discount and set $\gamma = 1$. Despite this formulation, the state $s_t$ is not fully observed, and is instead partially observed via an observation space $\mathcal{O} \in \mathbb{R}^{d_o}$. To allow us to apply standard RL techniques for MDPs, the agent is passed a recent trajectory of observations as a state representation i.e. $s_t = (o_t, o_{t-1},\ldots, o_{t-h})$ where $h$ is the history length \cite{kaelbling1998}. The agent selects actions using its policy $\pi: \mathcal{S} \rightarrow \mathcal{A}$, which it manipulates to maximise expected return from the current state $G_t = \mathbb{E} [\sum_{t=0}^H \gamma^t \mathcal{R}(s_t, a_t)]$, where $H$ is a finite time horizon.

We use the following sub-sections to detail the components of our proposed approach PEARL, and summarise our system in Figure \ref{fig:pearl} and Algorithm \ref{alg: pearl}.

\subsection{Prediction}
We follow the schema of model-based reinforcement learning where our task is to fit a function $\tilde{f}_\theta$ that approximates the true forward dynamics of the system $f(s_{t+1}, (s_t, a_t))$ given a dataset of experience collected from the environment $\mathcal{D} = \left[s_{n+1}, (s_n, a_n)\right]_{n=1}^N$. We employ probabilistic DNNs to learn this mapping, which provide data-efficient approximations of complex system dynamics and allow agents to incorporate prediction uncertainty into action selection \cite{gal2016, higuera2018}. Where traditional, \textit{deterministic} DNNs output point predictions given an input, here our probabilistic DNNs output distributions over the output nodes parameterised by a multivariate Gaussian distribution with mean $\boldsymbol{\mu}$ and diagonal covariance matrix $\Sigma$; i.e: $\tilde{f_\theta}(s_t, a_t) = \mathcal{N}(\mu_\theta(s_t, a_t), \Sigma_\theta(s_t, a_t))$. The agent maximises the likelihood of a target variable being drawn from the predicted distribution i.e. it performs Maximum Likelihood Estimation (MLE):
\begin{equation} \label{equation: loss}
    Loss(\theta) = \sum_{n=1}^N -\text{log}(P(s_n; \mu_\theta, \Sigma_\theta)) \; .
\end{equation}
By outputting a distribution over the next state our network can quantify \textit{aleotoric} uncertainty. Ensembling multiple probabilistic DNNs, training each on different subsets of the data, and averaging over their predictions can quantify \textit{epistemic} uncertainty \cite{lakshminarayanan2017}. Here, we employ $K$-many models, averaging the predictions to ensure both types of uncertainty are captured
\begin{equation}
    \tilde{f}_\theta(s_t, a_t) = \frac{1}{K} \sum_{k=1}^K \tilde{f}_{\theta_k} (s_t, a_t) \; .
\end{equation}
\subsection{Control} \label{subsection: control}
Between interactions with the environment, our agent plans by combining Model Predictive Path Integral (MPPI) \cite{williams2015} with Trajectory Sampling \cite{chua2018}--highlighted in {\color{BrickRed} red} in Algorithm \ref{alg: pearl}. Here, we sample action sequences $a_{t:t+H-1}^u \; \forall \; u \in U$ from our policy $\pi$, an initially arbitrary multivariate Gaussian distribution with diagonal covariance and parameters $(\mu^0, \sigma^0)_{t:t+H} \in \mathbb{R}^a$. Then, we duplicate $P$-many state-action pairs at the first planning timestep $(s_t^p, a_t) \; \forall \; p \in P$ called \textit{particles}. Each particle is assigned one bootstrap from the dynamics function ensemble and iteratively passed through it to create next-state distributions which can be sampled: $s_{t+1}^p \sim \mathcal{N}(\mu_{t+1}^p, \Sigma_{t+1}^p; \theta)$. We estimate the return $G_{\Gamma}$ from each trajectory $\Gamma$ by taking an expectation across particles
\begin{equation} \label{equation: expected reward}
    G_\Gamma = \mathbb{E}_{P} \left[\sum_{t=0}^H \mathcal{R}(s_t, a_t) \right] \; .
\end{equation}

We select the top-$e$ returns $G_\Gamma^*$, sometimes called the \textit{elite} sequences, and update the parameters of $\pi$ at iteration $j$ using a $G_\Gamma^*$-normalised estimate:
\begin{equation} \label{equation: mppi update}
\mu^j = \frac{\sum_{i=1}^e \Pi_i \Gamma_i^*}{\sum_{i=1}^e \Pi_i}, \; \sigma^j = \sqrt{\frac{\sum_{i=1}^e \Pi_i (\Gamma_i^* - \mu^j)^2}{\sum_{i=1}^e \Pi_i}} \; ,
\end{equation}

where $\Pi_i = e^{\tau(G_\Gamma^*, i)}$, $\tau$ is a temperature parameter that mediates weight given to the optimal trajectory, and $\Gamma_i^*$ is the $i$th top-$e$ trajectory corresponding to expected return $G_\Gamma^*$ \cite{hansen2022}. After $n$ optimisation iterations, $\mu^{n}$ represents the optimal action sequence, and the first action is taken in the real environment i.e. we perform receding horizon control.

\subsection{System Identification} \label{subsection: system ID}
We grant the agent a window, or \textit{commissioning period} $C$, to explore the state-action space and fit the one-step dynamics function $\tilde{f}_\theta(s_t, a_t)$. Our task is to sample the state-action space sequentially, and update the parameters $\theta$ of our model, such that we minimise the error in model predictions by the end of commissioning period $C$. Inspired by Bayesian Optimisation \cite{snoek2012}, we leverage the predictive variance in our models to select trajectories that transition the agent to parts of the state-space where it is most uncertain, sometimes called Maximum Variance (MV) exploration \cite{jain2018}. We adapt the routine from the previous section to evaluate the variance $V_\Gamma$ in the predicted rewards across particle trajectories 
\begin{equation}
    V_\Gamma = Var_P\left[\sum_{t=0}^H \mathcal{R}(s_t, a_t) \right] \; ,
\end{equation}
shown in {\color{OliveGreen} green} in Algorithm \ref{alg: pearl}. Now the elite trajectories $\Gamma^*$ in Equation \eqref{equation: mppi update} correspond to the action sequences and that maximise predictive variance $V_\Gamma^*$. We update the parameters of our model $\theta$ after each sample, thereafter we update $\theta$ at the end of each day.

\begin{algorithm}[tb]
   \caption{PEARL}
   \label{alg: pearl}
\begin{algorithmic}[1]
\REQUIRE $\mathcal{D}$, $\tilde{f}(s_t, a_t)_\theta$: memory, dynamics model \\
~~~~~~~~~~~$\mu^{0}, \sigma^{0}$, $N$: $\pi_{MPPI}$ params\\
~~~~~~~~~~~$\mathbf{s}_{t}, \textbf{a}_0$: current state, random init action\\
~~~~~~~~~~~$C, U, H$: comm. steps, act seqs., horizon\\
\FOR{step $t=0...T$}
\FOR{iteration $n=1...N$}
\FOR{act. seq. $u=1...U$}
\STATE{$a_{t:t+H-1}^u \sim \pi$} ~~~~~~~~~~~~~~~~~~~~~~~~{\color{CadetBlue}$\vartriangleleft$ \emph{Sample actions}}
\STATE {$s_{t+H}^p = \tilde{f}(s_t^p, a_{t:t+H-1}^i)_\theta \; \forall \; p \in P$} ~~~~~~~{\color{CadetBlue}$\vartriangleleft$ \emph{Plan}}
\IF{$t \leq C$}
\STATE {\color{OliveGreen}$V_\Gamma = Var_{P} \left[\sum_{t=0}^H \mathcal{R}(s_t, a_t) \right]$}
\STATE {\color{OliveGreen}$\mu^{n+1}, \sigma^{n+1}\leftarrow\mu^n(V_\Gamma), \sigma^n(V_\Gamma)$} ~~~~~~~~{\color{CadetBlue} $\vartriangleleft$ \emph{Eq. \ref{equation: mppi update}}}
\STATE {\color{OliveGreen} update $\tilde{f}(s_t, a_t)_\theta$ given $\mathcal{D}$} ~~~~~~~~~~~~~~~~~~{\color{CadetBlue}$\vartriangleleft$ \emph{Eq. \ref{equation: loss}}}
\ELSE
\STATE {\color{BrickRed} $G_\Gamma = \mathbb{E}_{P} \left[\sum_{t=0}^H \mathcal{R}(s_t, a_t) \right]$}
\STATE {\color{BrickRed} $\mu^{n+1}, \sigma^{n+1}\leftarrow\mu^n (G_\Gamma), \sigma^n(G_\Gamma)$} ~~~~~~~{\color{CadetBlue} $\vartriangleleft$ \emph{Eq. \ref{equation: mppi update}}}
\ENDIF
\ENDFOR
\STATE $a_{t} \leftarrow \mu_{t}^{N}$
\STATE $\mathcal{D} \leftarrow (s_t, a_t, r_t, s_{t+1})$
\ENDFOR
\IF{end of day}
\STATE {\color{BrickRed} update $\tilde{f}(s_t, a_t)_\theta$ given $\mathcal{D}$} ~~~~~~~~~~~~~~~~~~~~~~~~~~~~{\color{CadetBlue}$\vartriangleleft$ \emph{Eq. \ref{equation: loss}}}
\ENDIF
\ENDFOR
\end{algorithmic}
\end{algorithm}

\subsection{Reward Function}
The reward $\mathcal{R}(s_t, a_t)$ is a linear combination of an \textit{emissions term} $r_{E}[t]$ and a \textit{temperature term} $r_T[t]$ i.e. $\mathcal{R}(s_t, a_t) = r_{E}[t] + r_T[t]$ . Our goal is to motivate the agent to minimise emissions whilst satisfying thermal comfort in the building. If $r_{E}[t]$ is an emissions-term reward at timestep $t$, $E[t]$ is the total energy consumption in the environment at time $t$, and $C[t]$ the grid carbon intensity at time $t$, then the emissions-term reward is
\begin{equation}
	r_{E}[t] = - \phi \left(E[t] C[t] \right) \; ,
\end{equation}
where $\phi$ is a tunable parameter that sets the relative emphasis of emission-minimisation over thermal comfort (cf. Technical Appendix). The reward is negative because our goal is to minimise emissions, or maximise the negative of emissions produced. If $r_T^i[t]$ is a temperature-term reward at timestep $t$ for thermal zone $i$, $T_{obs}^i$ is the observed temperature in thermal zone $i$, and $T_{low}$ and $T_{high}$ are the lower and upper temperature bounds on thermal comfort respectively. The temperature reward is then given by
\begin{equation} \label{eq: temperature reward}
r_T^i[t]= 
\begin{cases}
		0:& T_{low} \leq T_{obs}^i \leq T_{high} \\
	     -\min [(T_{low} - T_{obs}^i[t])^2, \\
	     (T_{high} - T_{obs}^i)^2]: & \text{otherwise} \; , \\
\end{cases}
\end{equation}
where the second term can be thought of as a \textit{penalty} that punishes the agent in proportion to deviations from the thermal comfort zone. The total temperature reward $r_T[t]$ is obtained by summing the rewards across thermal zones i.e. $\sum_{i=1}^N r_T^i[t]$.

\section{Experimental Setup} \label{section: experiments}
\begin{figure*}[!ht]
    \centering
    \includegraphics[width=\textwidth]{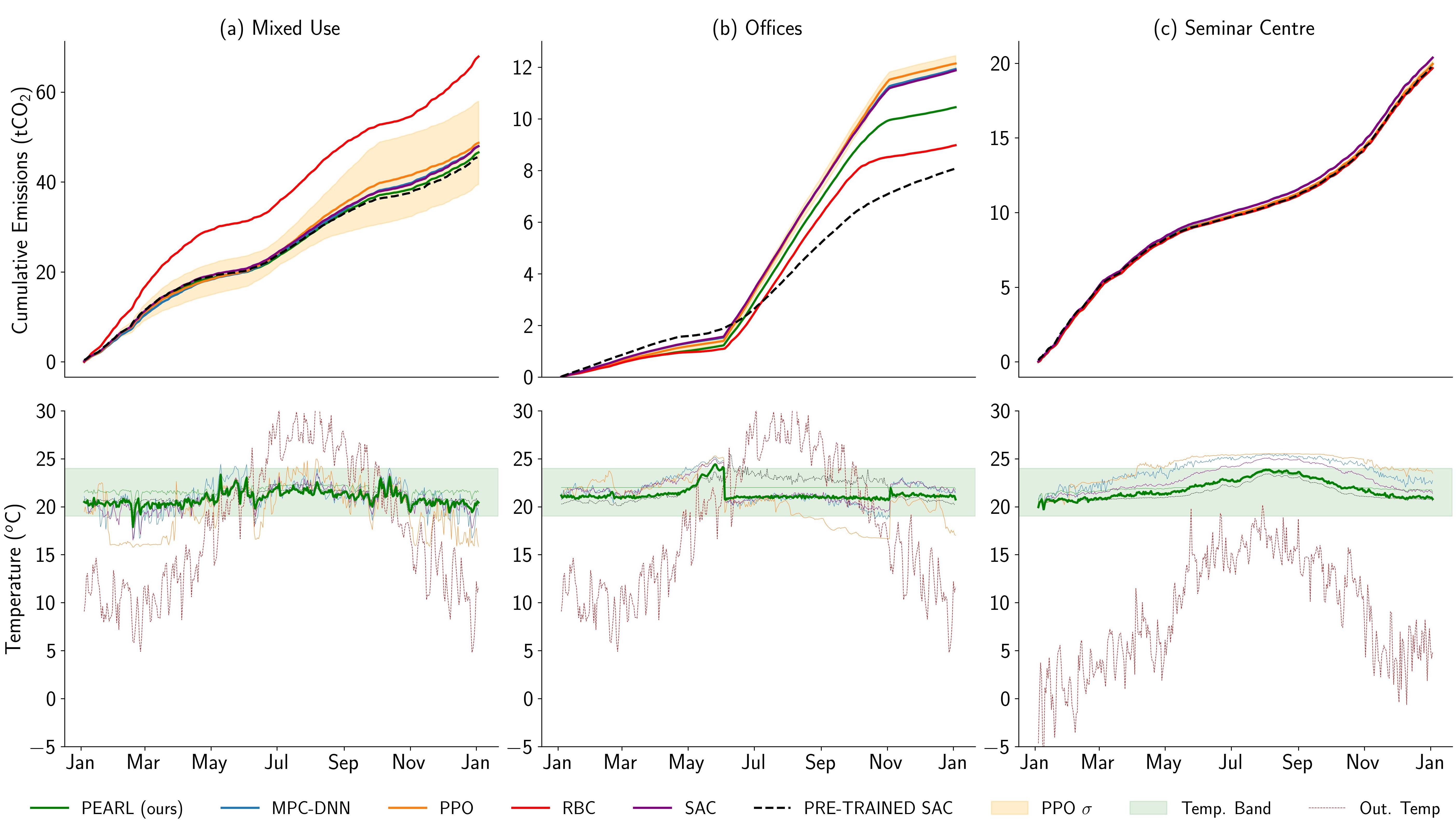}
    \caption{\textbf{\textit{Energym} Performance.} Top: Cumulative emissions produced by all agents across the (a) \textit{Mixed Use}, (b) \textit{Offices}, and (c) \textit{Seminar Centre} environments. Curves represent the mean of 3 runs of the experiment, shaded areas are one standard deviation (too small to see in all cases except PPO). Bottom: Mean daily building temperature produced by all agents, the green shaded area illustrates the target temperature range [19, 24].}
    \label{fig: emissions}
\end{figure*}

\begin{table*}[!ht] 
\centering
\caption{\label{table:results}\textbf{\textit{Energym} Performance.} Results for all agents across our three \textit{Energym} environments. We define the temperature infraction metric as the percentage of days where mean building temperature falls outside the target range [19, 24], and latency as the mean compute time each agent requires to select an action given its policy measured in seconds per action. Results are averaged across 3 runs and presented as mean $\pm$ standard deviation, except for the Oracle which has converged on a policy prior to deployment with multiple runs showing the same performance.}
\resizebox{0.8\textwidth}{!}{%
\begin{tabular}{llllll}
\toprule
                             & Agent   & Emissions (tCO$_2$) & Temp. Infractions &  Latency & Mean Reward \\
\midrule
\multirow{5}{*}{\rotatebox[]{90}{Mixed-Use}}   & RBC     &  $68.09 \scriptstyle{\pm0.00}$ & $2.47\% \scriptstyle{\pm0\%}$   & -- & --                   \\
                             & PPO     &   $48.80 \scriptstyle{\pm10.35}$  & $48.49\%\scriptstyle{\pm6.08\%}$ & $\textbf{0.021} \scriptstyle{\pm0.01}$  & $-1.47e7 \scriptstyle{\pm2.84e6}$ \\
                             & SAC & $48.80 \scriptstyle{\pm0.14}$ & $\mathbf{0\%}\scriptstyle{\pm0\%}$ &  $0.028 \scriptstyle{\pm0.02}$  & $-6.02e5 \scriptstyle{\pm3.85e4}$ \\
                             & MPC-DNN &    $48.03 \scriptstyle{\pm0.46}$ & $13.42\%\scriptstyle{\pm0.59\%}$ &  $0.030 \scriptstyle{\pm0.01}$ & $-1.12e6 \scriptstyle{\pm3.47e4}$ \\
                             & \textbf{PEARL} & $\mathbf{46.67 \scriptstyle{\pm0.09}}$ & $0.55\%\scriptstyle{\pm0.08\%}$ & $0.870 \scriptstyle{\pm 0.15}$ & $\mathbf{-5.76e10^5} \scriptstyle{\pm 2.12e10^3} $\\
                             \hline
                             & Oracle & $45.49$  & $0\%$ & $0.027$  &  $-4.77e5$ \\
\hline
\multirow{5}{*}{\rotatebox[]{90}{Office}}   & RBC     &    $\mathbf{9.61} \scriptstyle{\pm0.00}$ & $1.64\% \scriptstyle{\pm0\%}$& --     & --             \\
                             & PPO     & $12.14 \scriptstyle{\pm0.31}$ & $31.51\%\scriptstyle{\pm7.19\%}$ & $\mathbf{0.018} \scriptstyle{\pm0.006}$ & $-2.26e6 \scriptstyle{\pm 1.07e6}$ \\
                             & SAC & $11.87 \scriptstyle{\pm0.01}$ & $6.58\% \scriptstyle{\pm1.12\%}$  & $0.025 \scriptstyle{\pm0.01}$ & $-2.75e5 \scriptstyle{\pm 2.03e4}$ \\
                             & MPC-DNN & $11.93 \scriptstyle{\pm0.022}$ & $9.86\%\scriptstyle{\pm0.84\%}$ & $0.029 \scriptstyle{\pm0.001}$ & $-5.50e5 \scriptstyle{\pm2.89e4}$ \\
                             & \textbf{PEARL}   &  $10.45 \scriptstyle{\pm0.07}$ &  $\mathbf{1.52\%} \scriptstyle{\pm0\%}$ & $0.845 \scriptstyle{\pm 0.14}$ & $\mathbf{-5.51e10^4} \scriptstyle{\pm1.82e10^3}$ \\
                             \hline
                             & Oracle & $8.08$  & $2.47\%$ & $0.023$   &  $-2.63e4$ \\
\hline
\multirow{5}{*}{\rotatebox[]{90}{Sem. Centre}} & RBC     & $\mathbf{19.74} \scriptstyle{\pm0.00}$ &  $\mathbf{0\%} \scriptstyle{\pm0\%}$&  --   & --             \\
                             & PPO     &   $20.01 \scriptstyle{\pm0.27}$ &  $51.23\%\scriptstyle{\pm14.99\%}$  &  $\mathbf{0.028} \scriptstyle{\pm 0.01}$ & $-4.15e6 \scriptstyle{\pm9.67e5}$ \\
                             & SAC & $20.37 \scriptstyle{\pm0.08}$ & $29.32\%\scriptstyle{\pm1.87\%}$  & $0.033 \scriptstyle{\pm0.02}$  & $-1.95e6 \scriptstyle{\pm7.11e4}$ \\
                             & MPC-DNN &  $20.44 \scriptstyle{\pm0.02}$    & $49.13\%\scriptstyle{\pm0.56\%}$  & $0.035 \scriptstyle{\pm0.001}$  & $-2.45e6 \scriptstyle{\pm3.26e4}$ \\
                             & \textbf{PEARL}   &    $20.02 \scriptstyle{\pm0.10}$ & $\mathbf{0\%}\scriptstyle{\pm0\%}$  & $0.911 \scriptstyle{\pm 0.17}$  & $\mathbf{-1.18e10^6} \scriptstyle{\pm1.50e10^3}$ \\
                             \hline
                             & Oracle & $19.75$  & $0\%$ & $0.031$  &  $-1.14e6$ \\
\hline 
\end{tabular}}
\end{table*}

\subsection{Environments}
We evaluate the performance of our proposed approach using \textit{Energym}, an open-source building simulation library for benchmarking smart-grid control algorithms \cite{scharnhorst2021} that offers more candidate buildings that any other open-source package. \textit{Energym} provides a \textit{Python} interface for ground-truth building simulations designed in \textit{EnergyPlus} \cite{energyplus2001}, and presents buildings with varied equipment, geographies, and structural properties. We perform experiments in the following three buildings:
\begin{description}
    \item \textbf{\textit{Mixed-Use}} facility in Athens, Greece. 566.38m$^2$ surface area; 13 thermal zones; $\mathcal{A} \in \mathbb{R}^{12}$ and $\mathcal{S} \in \mathbb{R}^{37}$. Temperature setpoints and air handling unit (AHU) flowrates are controllable.
    \item \textbf{\textit{Office}} block in Athens, Greece. 643.73m$^2$ surface area; 25 thermal zones; $\mathcal{A} \in \mathbb{R}^{14}$ and $\mathcal{S} \in \mathbb{R}^{56}$. Only temperature setpoints are controllable.
    \item \textbf{\textit{Seminar Centre}} in Billund, Denmark. 1278.94m$^2$ surface area; 27 thermal zones; $\mathcal{A} \in \mathbb{R}^{18}$ and $\mathcal{S} \in \mathbb{R}^{59}$. Only temperature setpoints are controllable.
\end{description}
In all cases, environment states are represented by a combination of temperature, humidity and pressure sensors (among others). Full state and action spaces for each building are reported in the Technical Appendix for brevity. Experiments were run in each environment for one year starting on 01/01/2017,  advancing in $k$-minute timestep increments, with $k$ being environment dependent. Weather and grid carbon intensity match the true data in each geography for this period. 

\subsection{Baselines}
We compare the performance of our agent against several strong RL baselines, and an RBC:
\begin{description}
    \item \textbf{Soft Actor Critic} (SAC; \cite{Haarnoja2018}), a state-of-the-art algorithm, known for lower variance performance than other popular model-free algorithms like PPO and DDPG. 
    \item \textbf{Proximal Policy Optimisation} (PPO; \cite{schulman2017}), a popular model-free algorithm in production and used in previous works by \citet{ding2020} and \citet{zhang2019} as a baseline.
    \item \textbf{MPC with Deterministic Neural Networks} (MPC-DNN; \citet{nagabandi2018}), a simple, high-performing model-based architecture. Varying implementations have been used by previous authors, notably \citet{ding2020} and \citet{zhang2019}. We use the original implementation by \citet{nagabandi2018}.
    \item \textbf{RBC}, a generic, bang-bang controller found in most heating/cooling equipment that follows the heuristics outlined in the Technical Appendix.
    \item \textbf{Oracle}, an SAC agent with hyperparameters fit to each environment using Bayesian Optimsation in Weights and Biases \cite{wandb}, and \textbf{pre-trained} in each building simulation for 10 years prior to test time. 
\end{description}
We ensure both model-based agents plan with the same number of candidate actions over the same time horizon $H$ to ensure performance variation is a consequence only of their differing design. For each agent we adopt the implementations from their original papers, except for the number of network layers and their dimensions which are set to 5 and 200 respectively to attempt to capture the full complexity of the system dynamics. Full hyperparameter specifications are provided in the Technical Appendix--PEARL's trajectory sampling and MPPI hyperparameters follow implementations by \citet{chua2018} and \citet{hansen2022} respectively.

\section{Results and Discussion} \label{section: discussion}
\subsection{Performance} \label{subsection: performance results}

Table \ref{table:results} reports key metrics for our six controllers across our \textit{Energym} environments. Results are reported as the mean $\pm$ standard deviation for 3 runs of each experiment. The RBC performs identically across all experiments because both its policy and the environment are deterministic; varying RL agent performance is a consequence of policy and initialisation stochasticity.

\begin{figure}[t!]
    \centering
    \includegraphics[width= \columnwidth]{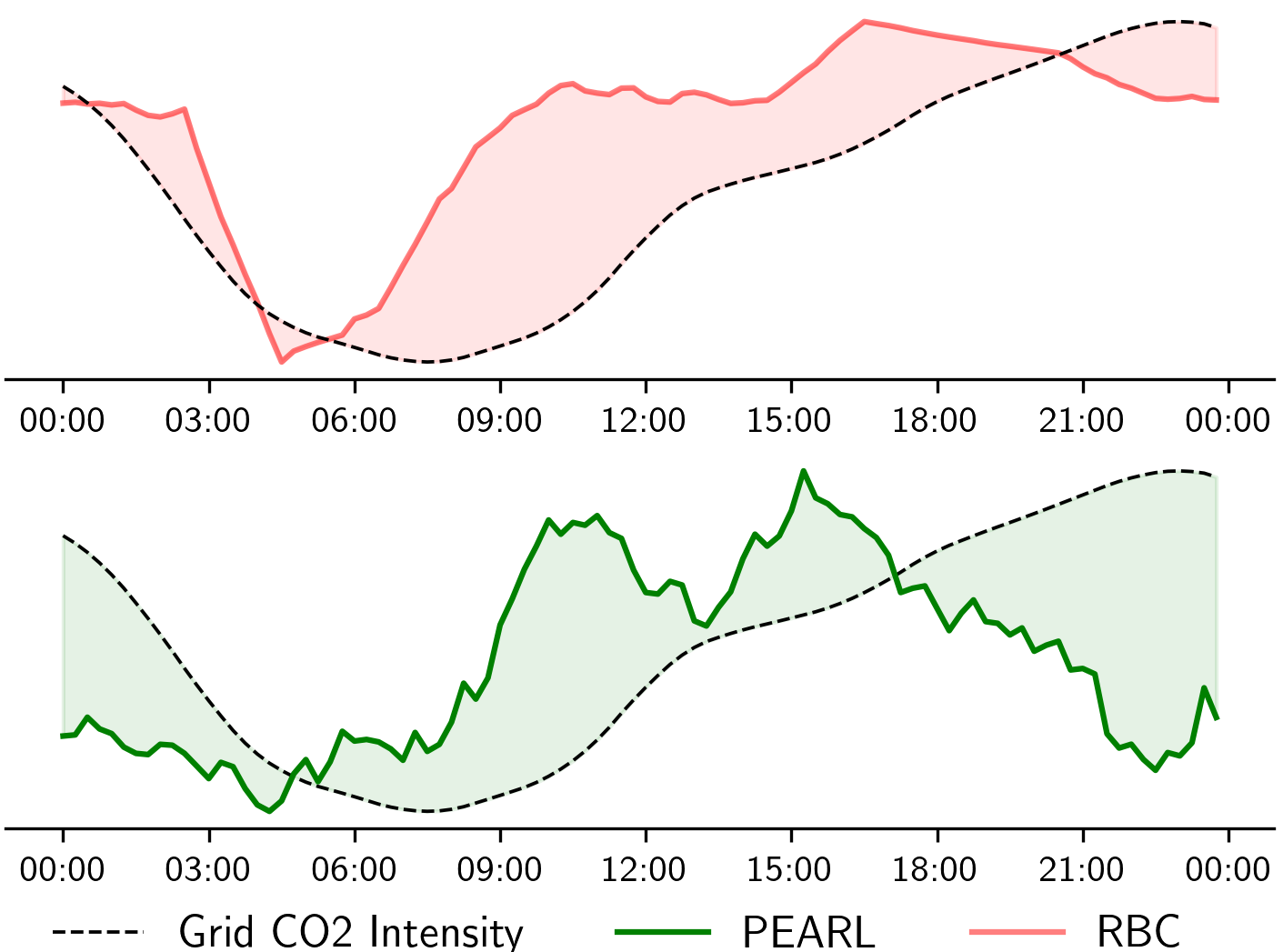}
    \caption{\textbf{Load Shifting.} Power consumption w.r.t. grid carbon intensity for the RBC (top) and PEARL (bottom) on an exemplar day in the \textit{Office} environment. We wish to maximise the shaded area to minimise emissions. PEARL minimises power draw in the early morning and late evening when grid carbon intensity is highest.}
    \label{fig:load-shifting}
\end{figure}

\textbf{Emissions.} We find PEARL produces minimum emissions in the \textit{Mixed-Use} environment, with cumulative emissions 31.46\% lower than the RBC. In the \textit{Office} block, the RBC exhibits lowest cumulative emissions, and PEARL outperforms all RL baselines. In the \textit{Seminar Centre}, the RBC minimises emissions, and PPO marginally outperforms PEARL, but does so at the cost of erroneous temperature control. Low external temperatures in the \textit{Seminar Centre} make experiments there less informative as the optimal policy is to heat most of the year, providing little room for improved control. Stepwise emission totals for each environment are illustrated in the top half of Figure \ref{fig: emissions}. We provide an illustrative example of PEARL showing an ability to load shift in Figure \ref{fig:load-shifting}.

\textbf{Temperature.} We find that PEARL produces minimum daily mean temperature infractions in the \textit{Office} and \textit{Seminar Centre} environments, and is slightly outperformed by SAC in the \textit{Mixed-Use} environment. The RBC is comparably performant across all environments, as would be expected. The remaining RL baselines miss the thermal bounds regularly, with PPO and MPC-DNN exhibiting temperature infraction rates as high as 51.23\% and 49.13\% respectively. In some cases the strong emissions performance of these baselines is a direct consequence of shutting off HVAC equipment and forcing uncomfortable internal temperatures. Mean daily building temperatures for each agent across the environments are plotted in the lower half of Figure \ref{fig: emissions}.

\textbf{Latency.} PPO is the lowest-latency (mean compute time per action) controller, selecting actions in two thirds of the time required by MPC-DNN, and 41 times faster than PEARL on average, but we note the latency of all agents is far smaller than the sampling period of each environment, meaning all implementations would prove adequate for real-world deployment. Were they deployed in situ, the model-based agents could utilise the time between environment interactions fully to plan with greater numbers of action sequences which we expect would improve performance.

\textbf{Reward.} Mean annual reward captures an agent's ability to minimise emissions \textit{and} maintain thermal comfort; this is the primary measure of agent performance. PEARL exhibits maximum mean reward across all environments suggesting it strikes this balance better than the other controllers. The reward curves for each agent are reported in the Technical Appendix for brevity.

\textbf{Oracle comparison.} The pre-trained oracle outperforms the baselines and PEARL in all cases as expected. However, its performance is surprisingly close to PEARL's, showing only 2.5\% and 1.3\% lower emissions in the \textit{Seminar Centre} and \textit{Mixed-Use} environments, and exhibiting similar thermal performance. From these results one could conclude that PEARL has produced a near-optimal policy, but one cannot be certain the oracle has reached optimality given the unusual, non-convergent reward curves this problem setting creates (cf. Technical Appendix). Indeed, the shape of the reward curve associated with an optimal policy is unclear, unlike the canonical episodic RL tasks (cartpole, mountain car etc.) where optimal solutions can be quantified by a fixed episodic return. Identifying optimality in this context is challenging and has been hitherto unexplored by the community.

\textbf{Why is \textit{Mixed-Use} performance an outlier?} An important observation from Table \ref{table:results} is that PEARL only outperforms the RBC in the \textit{Mixed-Use} facility. Why would this be the case? The outcome of the \textit{Seminar Centre} results has been discussed above, but one may expect PEARL to perform as well in the Office environment as it does in the \textit{Mixed-Use} environment. Unlike the Office environment, in the \textit{Mixed-Use} facility the agent has access to thermostat setpoints \textit{and} continuous AHU flowrate control. This greatly increases action-space complexity and moves the control problem away from a setting where simple heuristics can be readily applied. This would suggest that RL building controllers should only be deployed when the action space is sufficiently complex, likely owing to some access to continuous control parameters.
\begin{figure}[t!]
    \centering
    \includegraphics[width=\columnwidth]{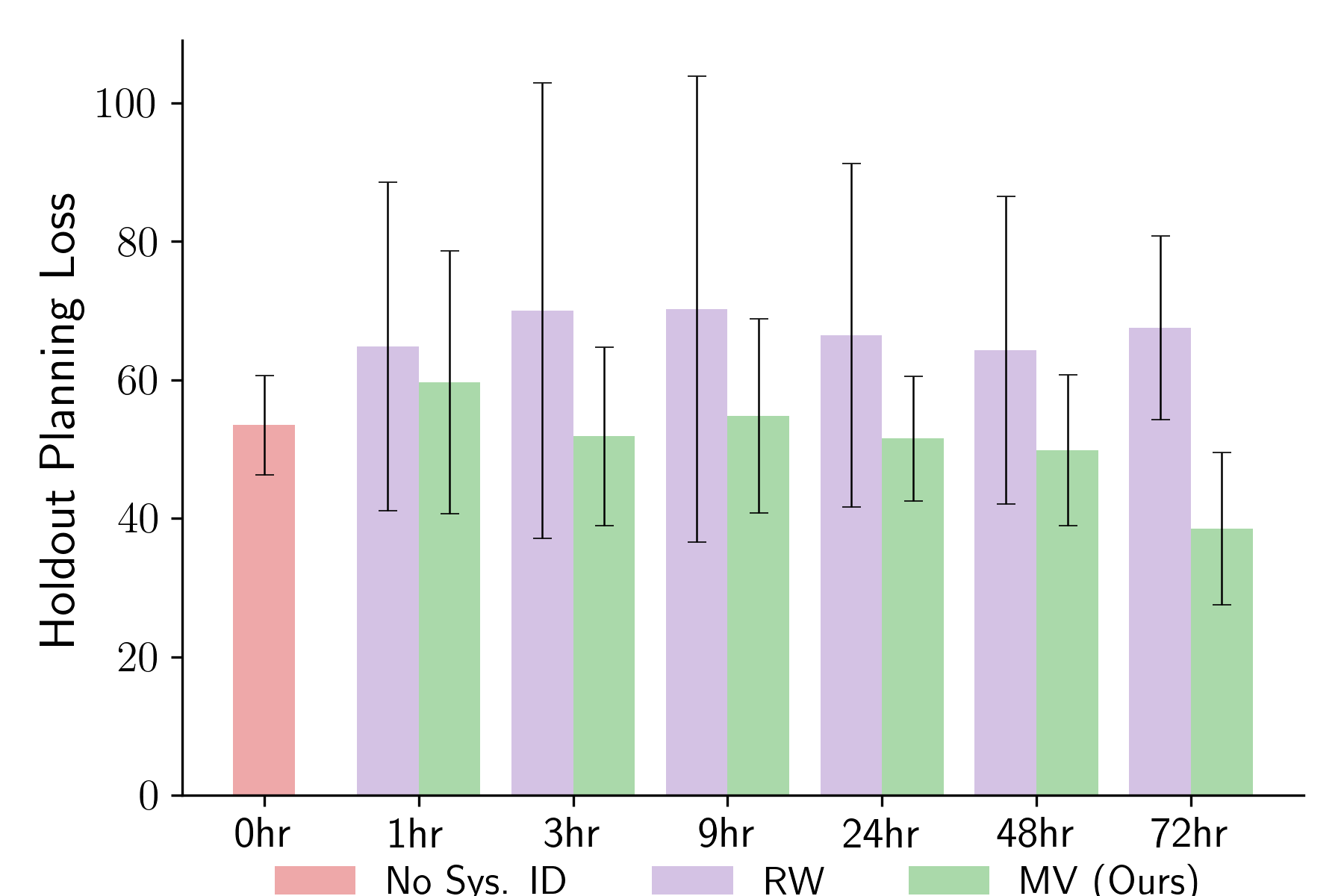}
    \caption{\textbf{System ID.} Planning MSE post-commissioning on a holdout set of 100 randomly sampled state-action trajectories, given varying system ID duration. Black bars represent one standard deviation across three runs.}
    \label{fig:system id}
\end{figure}

\subsection{System Identification}
We test the sensitivity of PEARL's performance to system ID duration. We vary the commissioning period at seven intervals between 0 (no system ID) and 72 hours, and test predictive accuracy on a holdout set of 100 randomly sampled state-action trajectories produced by another controller in the same environment. We compare our method (MV) with \citet{LAZIC2018}'s Random Walk (RW) and plot the results in Figure \ref{fig:system id}--see the Technical Appendix for \citet{LAZIC2018}'s exploration policy.

We find that the accuracy of models fitted via MV system ID correlates with commissioning period length, as would be expected. We observe that, in expectation, models fit using data collected via \citet{LAZIC2018}'s RW perform no better than randomly initialised networks. Using MV exploration, model accuracy is noticeably better than random only after 72 hours of system ID. However, we note from Figure \ref{fig: emissions}, that an agent with 3 hours of system ID time remains capable of reducing emissions and maintaining thermal comfort despite poorer predictive accuracy.

\subsection{Agent Decomposition}
What components of PEARL enable performant control? We note two differences between PEARL and our model-based RL baseline MPC-DNN: 1) The use of probabilistic networks, rather than deterministic networks, and 2) The use of MPPI planning, rather than random shooting (RS). To understand their relative importance, we vary the design of PEARL to either include or exclude these components and compare performance--Table \ref{table: agent decomposition}.
\begin{table}[!t] 
\centering
\caption{\label{table: agent decomposition}\textbf{Agent Decomposition.} Mean daily reward for four instantiations of PEARL varying the choice of network and planning algorithm. The \textit{Oracle} is reported as a baseline. Experiments conducted in the Mixed-Use environment for one year.}
\resizebox{0.8\columnwidth}{!}{%
\begin{tabular}{lll}
\toprule
                             & Deterministic   & Probabilistic  \\
\midrule
Random Shooting (RS) & $-7.11e5$ & $-7.30e5$   \\
MPPI & $\mathbf{-5.28e5}$  & $-5.65e5$   \\

\midrule
Oracle & $-4.77e5$
\end{tabular}
}
\end{table}

We find the performance of PEARL to be sensitive to the choice of planner and, to our surprise, insensitive to the choice of network, with agents composed of deterministic and probabilistic networks performing similarly when optimiser choice is controlled for. This is in contradiction to many works suggesting probabilistic modelling of dynamics function improves performance over deterministic models, particularly in complex, partially-observed state-action spaces like those exhibited in this study \cite{deisenroth2011pilco, chua2018, levine2018}. Given changes to training time between networks are insignificant, we continue to endorse probabilistic dynamics functions despite this result, as they may improve performance in settings beyond those used in this study.

\section{Conclusion}
In this work we consider the task of learning policies \textit{tabula rasa} that minimise emissions in buildings whilst ensuring thermal comfort, a considerably harder task than pre-training models in simulation before deployment. We have proposed PEARL (\textbf{P}robabilistic \textbf{E}mission-\textbf{A}bating \textbf{R}einforcement \textbf{L}earning), and shown it can reduce emissions from buildings by up to 31.46\% when compared with an RBC, by fitting polices \textit{online} without pre-training in simulation. When compared with existing RL baselines, our algorithm performs favourably, showing reduced emissions in all cases bar one, whilst maintaining thermal comfort more effectively. Our approach is simple to commission, requiring no historical data or simulator access \textit{a priori}, and capable of generalising across varied building archetypes. The scaled deployment of such systems could prove effective climate change mitigation tools.

\noindent \textbf{Acknowledgements.} We thank Arduin Findeis and Nantas Nardelli for helpful feedback. This work was supported by an EPSRC DTP Studentship (EP/T517847/1) and Emerson Electric.

\bibliography{bib}

\appendix
\onecolumn

\section{Energym Environments}\label{appendix: energym envs}
\pagenumbering{gobble}
		\begin{longtable}{c c c c c}
			\caption{ State-space $\mathcal{S}_{mixed}$ of the \textbf{\textit{Mixed-use} environment}. \label{table: mixed state space}} \\
			
			\textbf{Variable} & \textbf{Type} & \textbf{Lower Bound} & \textbf{Upper Bound} & \textbf{Description} \\
			\midrule
			\endfirsthead
			
			\multicolumn{5}{l}{{\tablename\  \thetable{} -- continued from previous page}} \\
			
			\textbf{Variable}     & \textbf{Type} & \textbf{Lower Bound} & \textbf{Upper Bound} & \textbf{Description}\\
			\midrule
			\endhead
			
			\hline
			\multicolumn{5}{r}{\textit{continued on next page}} \\
			\endfoot
			
			\hline \hline
			\endlastfoot
			
			{\color[HTML]{212529} Bd\_T\_AHU1}                 & {\color[HTML]{212529} scalar} & {\color[HTML]{212529} 10}      & {\color[HTML]{212529} 30}       & {\color[HTML]{212529} AHU 1 temperature (°C).}           \\
{\color[HTML]{212529} Bd\_T\_AHU2}                 & {\color[HTML]{212529} scalar} & {\color[HTML]{212529} 10}      & {\color[HTML]{212529} 30}       & {\color[HTML]{212529} AHU 2 temperature (°C).}           \\
{\color[HTML]{212529} Ext\_Irr}                    & {\color[HTML]{212529} scalar} & {\color[HTML]{212529} 0}       & {\color[HTML]{212529} 1000}     & {\color[HTML]{212529} Direct normal radiation (W/m2).}   \\
{\color[HTML]{212529} Ext\_P}                      & {\color[HTML]{212529} scalar} & {\color[HTML]{212529} 80000.0} & {\color[HTML]{212529} 130000.0} & {\color[HTML]{212529} Outdoor air pressure (Pa).}        \\
{\color[HTML]{212529} Ext\_RH}                     & {\color[HTML]{212529} scalar} & {\color[HTML]{212529} 0}       & {\color[HTML]{212529} 100}      & {\color[HTML]{212529} Outdoor realtive humidity (\%RH).} \\
{\color[HTML]{212529} Ext\_T}                      & {\color[HTML]{212529} scalar} & {\color[HTML]{212529} -10}     & {\color[HTML]{212529} 40}       & {\color[HTML]{212529} Outdoor temperature (°C).}         \\
{\color[HTML]{212529} Fa\_Pw\_All}                 & {\color[HTML]{212529} scalar} & {\color[HTML]{212529} 0}       & {\color[HTML]{212529} 50000.0}  & {\color[HTML]{212529} Total power consumtion (W).}       \\
{\color[HTML]{212529} Z02\_Fl\_Fan}                & {\color[HTML]{212529} scalar} & {\color[HTML]{212529} 0}       & {\color[HTML]{212529} 1}        & {\color[HTML]{212529} Zone 2 fan flow rate.}             \\
{\color[HTML]{212529} Z02\_RH}                     & {\color[HTML]{212529} scalar} & {\color[HTML]{212529} 0}       & {\color[HTML]{212529} 100}      & {\color[HTML]{212529} Zone 2 relative humidity (\%RH).}  \\
{\color[HTML]{212529} Z02\_T}                      & {\color[HTML]{212529} scalar} & {\color[HTML]{212529} 10}      & {\color[HTML]{212529} 40}       & {\color[HTML]{212529} Zone 2 temperature (°C).}          \\
{\color[HTML]{212529} Z02\_T\_Thermostat\_sp\_out} & {\color[HTML]{212529} scalar} & {\color[HTML]{212529} 16}      & {\color[HTML]{212529} 26}       & {\color[HTML]{212529} Zone 2 thermostat setpoint (°C).}  \\
{\color[HTML]{212529} Z03\_Fl\_Fan}                & {\color[HTML]{212529} scalar} & {\color[HTML]{212529} 0}       & {\color[HTML]{212529} 1}        & {\color[HTML]{212529} Zone 3 fan flow rate.}             \\
{\color[HTML]{212529} Z03\_RH}                     & {\color[HTML]{212529} scalar} & {\color[HTML]{212529} 0}       & {\color[HTML]{212529} 100}      & {\color[HTML]{212529} Zone 3 relative humidity (\%RH).}  \\
{\color[HTML]{212529} Z03\_T}                      & {\color[HTML]{212529} scalar} & {\color[HTML]{212529} 10}      & {\color[HTML]{212529} 40}       & {\color[HTML]{212529} Zone 3 temperature (°C).}          \\
{\color[HTML]{212529} Z03\_T\_Thermostat\_sp\_out} & {\color[HTML]{212529} scalar} & {\color[HTML]{212529} 16}      & {\color[HTML]{212529} 26}       & {\color[HTML]{212529} Zone 3 thermostat setpoint (°C).}  \\
{\color[HTML]{212529} Z04\_Fl\_Fan}                & {\color[HTML]{212529} scalar} & {\color[HTML]{212529} 0}       & {\color[HTML]{212529} 1}        & {\color[HTML]{212529} Zone 4 fan flow rate.}             \\
{\color[HTML]{212529} Z04\_RH}                     & {\color[HTML]{212529} scalar} & {\color[HTML]{212529} 0}       & {\color[HTML]{212529} 100}      & {\color[HTML]{212529} Zone 4 relative humidity (\%RH).}  \\
{\color[HTML]{212529} Z04\_T}                      & {\color[HTML]{212529} scalar} & {\color[HTML]{212529} 10}      & {\color[HTML]{212529} 40}       & {\color[HTML]{212529} Zone 4 temperature (°C).}          \\
{\color[HTML]{212529} Z04\_T\_Thermostat\_sp\_out} & {\color[HTML]{212529} scalar} & {\color[HTML]{212529} 16}      & {\color[HTML]{212529} 26}       & {\color[HTML]{212529} Zone 4 thermostat setpoint (°C).}  \\
{\color[HTML]{212529} Z05\_RH}                     & {\color[HTML]{212529} scalar} & {\color[HTML]{212529} 0}       & {\color[HTML]{212529} 100}      & {\color[HTML]{212529} Zone 5 relative humidity (\%RH).}  \\
{\color[HTML]{212529} Z05\_T}                      & {\color[HTML]{212529} scalar} & {\color[HTML]{212529} 10}      & {\color[HTML]{212529} 40}       & {\color[HTML]{212529} Zone 5 temperature (°C).}          \\
{\color[HTML]{212529} Z05\_T\_Thermostat\_sp\_out} & {\color[HTML]{212529} scalar} & {\color[HTML]{212529} 16}      & {\color[HTML]{212529} 26}       & {\color[HTML]{212529} Zone 5 thermostat setpoint (°C).}  \\
{\color[HTML]{212529} Z08\_Fl\_Fan}                & {\color[HTML]{212529} scalar} & {\color[HTML]{212529} 0}       & {\color[HTML]{212529} 1}        & {\color[HTML]{212529} Zone 8 fan flow rate.}             \\
{\color[HTML]{212529} Z08\_RH}                     & {\color[HTML]{212529} scalar} & {\color[HTML]{212529} 0}       & {\color[HTML]{212529} 100}      & {\color[HTML]{212529} Zone 8 relative humidity (\%RH).}  \\
{\color[HTML]{212529} Z08\_T}                      & {\color[HTML]{212529} scalar} & {\color[HTML]{212529} 10}      & {\color[HTML]{212529} 40}       & {\color[HTML]{212529} Zone 8 temperature (°C).}          \\
{\color[HTML]{212529} Z08\_T\_Thermostat\_sp\_out} & {\color[HTML]{212529} scalar} & {\color[HTML]{212529} 16}      & {\color[HTML]{212529} 26}       & {\color[HTML]{212529} Zone 8 thermostat setpoint (°C).}  \\
{\color[HTML]{212529} Z09\_Fl\_Fan}                & {\color[HTML]{212529} scalar} & {\color[HTML]{212529} 0}       & {\color[HTML]{212529} 1}        & {\color[HTML]{212529} Zone 9 fan flow rate.}             \\
{\color[HTML]{212529} Z09\_RH}                     & {\color[HTML]{212529} scalar} & {\color[HTML]{212529} 0}       & {\color[HTML]{212529} 100}      & {\color[HTML]{212529} Zone 9 relative humidity (\%RH).}  \\
{\color[HTML]{212529} Z09\_T}                      & {\color[HTML]{212529} scalar} & {\color[HTML]{212529} 10}      & {\color[HTML]{212529} 40}       & {\color[HTML]{212529} Zone 9 temperature (°C).}          \\
{\color[HTML]{212529} Z09\_T\_Thermostat\_sp\_out} & {\color[HTML]{212529} scalar} & {\color[HTML]{212529} 16}      & {\color[HTML]{212529} 26}       & {\color[HTML]{212529} Zone 9 thermostat setpoint (°C).}  \\
{\color[HTML]{212529} Z10\_Fl\_Fan}                & {\color[HTML]{212529} scalar} & {\color[HTML]{212529} 0}       & {\color[HTML]{212529} 1}        & {\color[HTML]{212529} Zone 10 fan flow rate.}            \\
{\color[HTML]{212529} Z10\_RH}                     & {\color[HTML]{212529} scalar} & {\color[HTML]{212529} 0}       & {\color[HTML]{212529} 100}      & {\color[HTML]{212529} Zone 10 relative humidity (\%RH).} \\
{\color[HTML]{212529} Z10\_T}                      & {\color[HTML]{212529} scalar} & {\color[HTML]{212529} 10}      & {\color[HTML]{212529} 40}       & {\color[HTML]{212529} Zone 10 temperature (°C).}         \\
{\color[HTML]{212529} Z10\_T\_Thermostat\_sp\_out} & {\color[HTML]{212529} scalar} & {\color[HTML]{212529} 16}      & {\color[HTML]{212529} 26}       & {\color[HTML]{212529} Zone 10 thermostat setpoint (°C).} \\
{\color[HTML]{212529} Z11\_Fl\_Fan}                & {\color[HTML]{212529} scalar} & {\color[HTML]{212529} 0}       & {\color[HTML]{212529} 1}        & {\color[HTML]{212529} Zone 11 fan flow rate.}            \\
{\color[HTML]{212529} Z11\_RH}                     & {\color[HTML]{212529} scalar} & {\color[HTML]{212529} 0}       & {\color[HTML]{212529} 100}      & {\color[HTML]{212529} Zone 11 relative humidity (\%RH).} \\
{\color[HTML]{212529} Z11\_T}                      & {\color[HTML]{212529} scalar} & {\color[HTML]{212529} 10}      & {\color[HTML]{212529} 40}       & {\color[HTML]{212529} Zone 11 temperature (°C).}         \\
{\color[HTML]{212529} Z11\_T\_Thermostat\_sp\_out} & {\color[HTML]{212529} scalar} & {\color[HTML]{212529} 16}      & {\color[HTML]{212529} 26}       & {\color[HTML]{212529} Zone 11 thermostat setpoint (°C).} \\
{\color[HTML]{212529} Grid\_CO2}                   & {\color[HTML]{212529} scalar} & {\color[HTML]{212529} 0}     & {\color[HTML]{212529} 1000}     & {\color[HTML]{212529} Grid CO2 emission intensity (g/kWh).}
			\end{longtable}

	\begin{longtable}[c]{ c  c  c  c  c }
			\caption{\label{table: office state space} State-space \(\mathcal{S}_{office}\) of the \textbf{\textit{Offices} environment.}} \\
			
			\textbf{Variable}     & \textbf{Type} & \textbf{Lower Bound} & \textbf{Upper Bound} & \textbf{Description}\\
			\midrule
			\endfirsthead
			
			\multicolumn{5}{l}{{\tablename\  \thetable{} -- continued from previous page}} \\
			
			\textbf{Variable}     & \textbf{Type} & \textbf{Lower Bound} & \textbf{Upper Bound} & \textbf{Description}\\
			\midrule
			\endhead
			
			\hline
			\multicolumn{5}{r}{\textit{continued on next page}}
			\endfoot
			
			\hline \hline
			\endlastfoot
			{\color[HTML]{212529} Bd\_Pw\_All}                 & {\color[HTML]{212529} scalar} & {\color[HTML]{212529} 0}   & {\color[HTML]{212529} 5000}    & {\color[HTML]{212529} Building power consumption (W).}   \\
{\color[HTML]{212529} Ext\_Irr}                    & {\color[HTML]{212529} scalar} & {\color[HTML]{212529} 0}   & {\color[HTML]{212529} 1000}    & {\color[HTML]{212529} Direct normal radiation (W/m2).}   \\
{\color[HTML]{212529} Ext\_RH}                     & {\color[HTML]{212529} scalar} & {\color[HTML]{212529} 0}   & {\color[HTML]{212529} 100}     & {\color[HTML]{212529} Outdoor relative humidity (\%RH).} \\
{\color[HTML]{212529} Ext\_T}                      & {\color[HTML]{212529} scalar} & {\color[HTML]{212529} -10} & {\color[HTML]{212529} 40}      & {\color[HTML]{212529} Outdoor temperature (°C).}         \\
{\color[HTML]{212529} Fa\_Pw\_All}                 & {\color[HTML]{212529} scalar} & {\color[HTML]{212529} 0}   & {\color[HTML]{212529} 10000.0} & {\color[HTML]{212529} Total power consumption (W).}      \\
{\color[HTML]{212529} Fa\_Pw\_HVAC}                & {\color[HTML]{212529} scalar} & {\color[HTML]{212529} 0}   & {\color[HTML]{212529} 10000.0} & {\color[HTML]{212529} HVAC power consumption (W).}       \\
{\color[HTML]{212529} Fa\_Pw\_PV}                  & {\color[HTML]{212529} scalar} & {\color[HTML]{212529} 0}   & {\color[HTML]{212529} 2000.0}  & {\color[HTML]{212529} PV power production (W).}          \\
{\color[HTML]{212529} Z01\_Fl\_Fan\_sp\_out}       & {\color[HTML]{212529} scalar} & {\color[HTML]{212529} 0}   & {\color[HTML]{212529} 1}       & {\color[HTML]{212529} Zone 1 fan flow setpoint.}         \\
{\color[HTML]{212529} Z01\_T}                      & {\color[HTML]{212529} scalar} & {\color[HTML]{212529} 10}  & {\color[HTML]{212529} 40}      & {\color[HTML]{212529} Zone 1 temperature (°C).}          \\
{\color[HTML]{212529} Z01\_T\_Thermostat\_sp\_out} & {\color[HTML]{212529} scalar} & {\color[HTML]{212529} 16}  & {\color[HTML]{212529} 26}      & {\color[HTML]{212529} Zone 1 thermostat setpoint (°C).}  \\
{\color[HTML]{212529} Z02\_Fl\_Fan\_sp\_out}       & {\color[HTML]{212529} scalar} & {\color[HTML]{212529} 0}   & {\color[HTML]{212529} 1}       & {\color[HTML]{212529} Zone 2 fan flow setpoint.}         \\
{\color[HTML]{212529} Z02\_T}                      & {\color[HTML]{212529} scalar} & {\color[HTML]{212529} 10}  & {\color[HTML]{212529} 40}      & {\color[HTML]{212529} Zone 2 temperature (°C).}          \\
{\color[HTML]{212529} Z02\_T\_Thermostat\_sp\_out} & {\color[HTML]{212529} scalar} & {\color[HTML]{212529} 16}  & {\color[HTML]{212529} 26}      & {\color[HTML]{212529} Zone 2 thermostat setpoint (°C).}  \\
{\color[HTML]{212529} Z03\_Fl\_Fan\_sp\_out}       & {\color[HTML]{212529} scalar} & {\color[HTML]{212529} 0}   & {\color[HTML]{212529} 1}       & {\color[HTML]{212529} Zone 3 fan flow setpoint.}         \\
{\color[HTML]{212529} Z03\_Fl\_Fan1\_sp\_out}      & {\color[HTML]{212529} scalar} & {\color[HTML]{212529} 0}   & {\color[HTML]{212529} 1}       & {\color[HTML]{212529} Zone 3 fan 1 flow setpoint.}       \\
{\color[HTML]{212529} Z03\_T}                      & {\color[HTML]{212529} scalar} & {\color[HTML]{212529} 10}  & {\color[HTML]{212529} 40}      & {\color[HTML]{212529} Zone 3 temperature (°C).}          \\
{\color[HTML]{212529} Z03\_T\_Thermostat\_sp\_out} & {\color[HTML]{212529} scalar} & {\color[HTML]{212529} 16}  & {\color[HTML]{212529} 26}      & {\color[HTML]{212529} Zone 3 thermostat setpoint (°C).}  \\
{\color[HTML]{212529} Z04\_Fl\_Fan\_sp\_out}       & {\color[HTML]{212529} scalar} & {\color[HTML]{212529} 0}   & {\color[HTML]{212529} 1}       & {\color[HTML]{212529} Zone 4 fan flow setpoint.}         \\
{\color[HTML]{212529} Z04\_T}                      & {\color[HTML]{212529} scalar} & {\color[HTML]{212529} 10}  & {\color[HTML]{212529} 40}      & {\color[HTML]{212529} Zone 4 temperature (°C).}          \\
{\color[HTML]{212529} Z04\_T\_Thermostat\_sp\_out} & {\color[HTML]{212529} scalar} & {\color[HTML]{212529} 16}  & {\color[HTML]{212529} 26}      & {\color[HTML]{212529} Zone 4 thermostat setpoint (°C).}  \\
{\color[HTML]{212529} Z05\_Fl\_Fan\_sp\_out}       & {\color[HTML]{212529} scalar} & {\color[HTML]{212529} 0}   & {\color[HTML]{212529} 1}       & {\color[HTML]{212529} Zone 5 fan flow setpoint.}         \\
{\color[HTML]{212529} Z05\_T}                      & {\color[HTML]{212529} scalar} & {\color[HTML]{212529} 10}  & {\color[HTML]{212529} 40}      & {\color[HTML]{212529} Zone 5 temperature (°C).}          \\
{\color[HTML]{212529} Z05\_T\_Thermostat\_sp\_out} & {\color[HTML]{212529} scalar} & {\color[HTML]{212529} 16}  & {\color[HTML]{212529} 26}      & {\color[HTML]{212529} Zone 5 thermostat setpoint (°C).}  \\
{\color[HTML]{212529} Z06\_Fl\_Fan\_sp\_out}       & {\color[HTML]{212529} scalar} & {\color[HTML]{212529} 0}   & {\color[HTML]{212529} 1}       & {\color[HTML]{212529} Zone 6 fan flow setpoint.}         \\
{\color[HTML]{212529} Z06\_T}                      & {\color[HTML]{212529} scalar} & {\color[HTML]{212529} 10}  & {\color[HTML]{212529} 40}      & {\color[HTML]{212529} Zone 6 temperature (°C).}          \\
{\color[HTML]{212529} Z06\_T\_Thermostat\_sp\_out} & {\color[HTML]{212529} scalar} & {\color[HTML]{212529} 16}  & {\color[HTML]{212529} 26}      & {\color[HTML]{212529} Zone 6 thermostat setpoint (°C).}  \\
{\color[HTML]{212529} Z07\_Fl\_Fan\_sp\_out}       & {\color[HTML]{212529} scalar} & {\color[HTML]{212529} 0}   & {\color[HTML]{212529} 1}       & {\color[HTML]{212529} Zone 7 fan flow setpoint.}         \\
{\color[HTML]{212529} Z07\_T}                      & {\color[HTML]{212529} scalar} & {\color[HTML]{212529} 10}  & {\color[HTML]{212529} 40}      & {\color[HTML]{212529} Zone 7 temperature (°C).}          \\
{\color[HTML]{212529} Z07\_T\_Thermostat\_sp\_out} & {\color[HTML]{212529} scalar} & {\color[HTML]{212529} 16}  & {\color[HTML]{212529} 26}      & {\color[HTML]{212529} Zone 7 thermostat setpoint (°C).}  \\
{\color[HTML]{212529} Z15\_Fl\_Fan\_sp\_out}       & {\color[HTML]{212529} scalar} & {\color[HTML]{212529} 0}   & {\color[HTML]{212529} 1}       & {\color[HTML]{212529} Zone 15 fan flow setpoint.}        \\
{\color[HTML]{212529} Z15\_T}                      & {\color[HTML]{212529} scalar} & {\color[HTML]{212529} 10}  & {\color[HTML]{212529} 40}      & {\color[HTML]{212529} Zone 15 temperature (°C).}         \\
{\color[HTML]{212529} Z15\_T\_Thermostat\_sp\_out} & {\color[HTML]{212529} scalar} & {\color[HTML]{212529} 16}  & {\color[HTML]{212529} 26}      & {\color[HTML]{212529} Zone 15 thermostat setpoint (°C).} \\
{\color[HTML]{212529} Z16\_Fl\_Fan\_sp\_out}       & {\color[HTML]{212529} scalar} & {\color[HTML]{212529} 0}   & {\color[HTML]{212529} 1}       & {\color[HTML]{212529} Zone 16 fan flow setpoint.}        \\
{\color[HTML]{212529} Z16\_T}                      & {\color[HTML]{212529} scalar} & {\color[HTML]{212529} 10}  & {\color[HTML]{212529} 40}      & {\color[HTML]{212529} Zone 16 temperature (°C).}         \\
{\color[HTML]{212529} Z16\_T\_Thermostat\_sp\_out} & {\color[HTML]{212529} scalar} & {\color[HTML]{212529} 16}  & {\color[HTML]{212529} 26}      & {\color[HTML]{212529} Zone 16 thermostat setpoint (°C).} \\
{\color[HTML]{212529} Z17\_Fl\_Fan\_sp\_out}       & {\color[HTML]{212529} scalar} & {\color[HTML]{212529} 0}   & {\color[HTML]{212529} 1}       & {\color[HTML]{212529} Zone 17 fan flow setpoint.}        \\
{\color[HTML]{212529} Z17\_Fl\_Fan1\_sp\_out}      & {\color[HTML]{212529} scalar} & {\color[HTML]{212529} 0}   & {\color[HTML]{212529} 1}       & {\color[HTML]{212529} Zone 17 fan 1 flow setpoint.}      \\
{\color[HTML]{212529} Z17\_T}                      & {\color[HTML]{212529} scalar} & {\color[HTML]{212529} 10}  & {\color[HTML]{212529} 40}      & {\color[HTML]{212529} Zone 17 temperature (°C).}         \\
{\color[HTML]{212529} Z17\_T\_Thermostat\_sp\_out} & {\color[HTML]{212529} scalar} & {\color[HTML]{212529} 16}  & {\color[HTML]{212529} 26}      & {\color[HTML]{212529} Zone 17 thermostat setpoint (°C).} \\
{\color[HTML]{212529} Z18\_Fl\_Fan\_sp\_out}       & {\color[HTML]{212529} scalar} & {\color[HTML]{212529} 0}   & {\color[HTML]{212529} 1}       & {\color[HTML]{212529} Zone 18 fan flow setpoint.}        \\
{\color[HTML]{212529} Z18\_T}                      & {\color[HTML]{212529} scalar} & {\color[HTML]{212529} 10}  & {\color[HTML]{212529} 40}      & {\color[HTML]{212529} Zone 18 temperature (°C).}         \\
{\color[HTML]{212529} Z18\_T\_Thermostat\_sp\_out} & {\color[HTML]{212529} scalar} & {\color[HTML]{212529} 16}  & {\color[HTML]{212529} 26}      & {\color[HTML]{212529} Zone 18 thermostat setpoint (°C).} \\
{\color[HTML]{212529} Z19\_Fl\_Fan\_sp\_out}       & {\color[HTML]{212529} scalar} & {\color[HTML]{212529} 0}   & {\color[HTML]{212529} 1}       & {\color[HTML]{212529} Zone 19 fan flow setpoint.}        \\
{\color[HTML]{212529} Z19\_Fl\_Fan1\_sp\_out}      & {\color[HTML]{212529} scalar} & {\color[HTML]{212529} 0}   & {\color[HTML]{212529} 1}       & {\color[HTML]{212529} Zone 19 fan 1 flow setpoint.}      \\
{\color[HTML]{212529} Z19\_T}                      & {\color[HTML]{212529} scalar} & {\color[HTML]{212529} 10}  & {\color[HTML]{212529} 40}      & {\color[HTML]{212529} Zone 19 temperature (°C).}         \\
{\color[HTML]{212529} Z19\_T\_Thermostat\_sp\_out} & {\color[HTML]{212529} scalar} & {\color[HTML]{212529} 16}  & {\color[HTML]{212529} 26}      & {\color[HTML]{212529} Zone 19 thermostat setpoint (°C).} \\
{\color[HTML]{212529} Z20\_Fl\_Fan\_sp\_out}       & {\color[HTML]{212529} scalar} & {\color[HTML]{212529} 0}   & {\color[HTML]{212529} 1}       & {\color[HTML]{212529} Zone 20 fan flow setpoint.}        \\
{\color[HTML]{212529} Z20\_Fl\_Fan1\_sp\_out}      & {\color[HTML]{212529} scalar} & {\color[HTML]{212529} 0}   & {\color[HTML]{212529} 1}       & {\color[HTML]{212529} Zone 20 fan 1 flow setpoint.}      \\
{\color[HTML]{212529} Z20\_T}                      & {\color[HTML]{212529} scalar} & {\color[HTML]{212529} 10}  & {\color[HTML]{212529} 40}      & {\color[HTML]{212529} Zone 20 temperature (°C).}         \\
{\color[HTML]{212529} Z20\_T\_Thermostat\_sp\_out} & {\color[HTML]{212529} scalar} & {\color[HTML]{212529} 16}  & {\color[HTML]{212529} 26}      & {\color[HTML]{212529} Zone 20 thermostat setpoint (°C).} \\
{\color[HTML]{212529} Z25\_Fl\_Fan\_sp\_out}       & {\color[HTML]{212529} scalar} & {\color[HTML]{212529} 0}   & {\color[HTML]{212529} 1}       & {\color[HTML]{212529} Zone 25 fan flow setpoint.}        \\
{\color[HTML]{212529} Z25\_Fl\_Fan1\_sp\_out}      & {\color[HTML]{212529} scalar} & {\color[HTML]{212529} 0}   & {\color[HTML]{212529} 1}       & {\color[HTML]{212529} Zone 25 fan 1 flow setpoint.}      \\
{\color[HTML]{212529} Z25\_T}                      & {\color[HTML]{212529} scalar} & {\color[HTML]{212529} 10}  & {\color[HTML]{212529} 40}      & {\color[HTML]{212529} Zone 25 temperature (°C).}         \\
{\color[HTML]{212529} Z25\_T\_Thermostat\_sp\_out} & {\color[HTML]{212529} scalar} & {\color[HTML]{212529} 16}  & {\color[HTML]{212529} 26}      & {\color[HTML]{212529} Zone 25 thermostat setpoint (°C).} \\
{\color[HTML]{212529} Grid\_CO2}                   & {\color[HTML]{212529} scalar} & {\color[HTML]{212529} 0}     & {\color[HTML]{212529} 1000}     & {\color[HTML]{212529} Grid CO2 emission intensity (g/kWh).}
						\end{longtable}
						
			\begin{longtable}[c]{ c  c  c  c  c }
			\caption{\label{table: office action space} Action-space \(\mathcal{A}_{offices}\) of the \textbf{\textit{Offices} environment.}} \\
			
			\textbf{Action}     & \textbf{Type} & \textbf{Lower Bound} & \textbf{Upper Bound} & \textbf{Description}\\
			\midrule
			\endfirsthead
			
			\multicolumn{5}{l}{{\tablename\  \thetable{} -- continued from previous page}} \\
			
			\textbf{Action}     & \textbf{Type} & \textbf{Lower Bound} & \textbf{Upper Bound} & \textbf{Description}\\
			\midrule
			\endhead
			
			\hline
			\multicolumn{5}{r}{\textit{continued on next page}}
			\endfoot
			
			\hline \hline
			\endlastfoot

{\color[HTML]{212529} Z01\_T\_Thermostat\_sp} & {\color[HTML]{212529} scalar}   & {\color[HTML]{212529} 16} & {\color[HTML]{212529} 26} & {\color[HTML]{212529} Zone 1 thermostat setpoint (°C).}      \\

{\color[HTML]{212529} Z02\_T\_Thermostat\_sp} & {\color[HTML]{212529} scalar}   & {\color[HTML]{212529} 16} & {\color[HTML]{212529} 26} & {\color[HTML]{212529} Zone 2 thermostat setpoint (°C).}      \\

{\color[HTML]{212529} Z03\_T\_Thermostat\_sp} & {\color[HTML]{212529} scalar}   & {\color[HTML]{212529} 16} & {\color[HTML]{212529} 26} & {\color[HTML]{212529} Zone 3 thermostat setpoint (°C).}      \\

{\color[HTML]{212529} Z04\_T\_Thermostat\_sp} & {\color[HTML]{212529} scalar}   & {\color[HTML]{212529} 16} & {\color[HTML]{212529} 26} & {\color[HTML]{212529} Zone 4 thermostat setpoint (°C).}      \\

{\color[HTML]{212529} Z05\_T\_Thermostat\_sp} & {\color[HTML]{212529} scalar}   & {\color[HTML]{212529} 16} & {\color[HTML]{212529} 26} & {\color[HTML]{212529} Zone 5 thermostat setpoint (°C).}      \\

{\color[HTML]{212529} Z06\_T\_Thermostat\_sp} & {\color[HTML]{212529} scalar}   & {\color[HTML]{212529} 16} & {\color[HTML]{212529} 26} & {\color[HTML]{212529} Zone 6 thermostat setpoint (°C).}      \\

{\color[HTML]{212529} Z07\_T\_Thermostat\_sp} & {\color[HTML]{212529} scalar}   & {\color[HTML]{212529} 16} & {\color[HTML]{212529} 26} & {\color[HTML]{212529} Zone 7 thermostat setpoint (°C).}      \\

{\color[HTML]{212529} Z15\_T\_Thermostat\_sp} & {\color[HTML]{212529} scalar}   & {\color[HTML]{212529} 16} & {\color[HTML]{212529} 26} & {\color[HTML]{212529} Zone 15 thermostat setpoint (°C).}     \\

{\color[HTML]{212529} Z16\_T\_Thermostat\_sp} & {\color[HTML]{212529} scalar}   & {\color[HTML]{212529} 16} & {\color[HTML]{212529} 26} & {\color[HTML]{212529} Zone 16 thermostat setpoint (°C).}     \\
 
{\color[HTML]{212529} Z17\_T\_Thermostat\_sp} & {\color[HTML]{212529} scalar}   & {\color[HTML]{212529} 16} & {\color[HTML]{212529} 26} & {\color[HTML]{212529} Zone 17 thermostat setpoint (°C).}     \\
 
{\color[HTML]{212529} Z18\_T\_Thermostat\_sp} & {\color[HTML]{212529} scalar}   & {\color[HTML]{212529} 16} & {\color[HTML]{212529} 26} & {\color[HTML]{212529} Zone 18 thermostat setpoint (°C).}     \\

{\color[HTML]{212529} Z19\_T\_Thermostat\_sp} & {\color[HTML]{212529} scalar}   & {\color[HTML]{212529} 16} & {\color[HTML]{212529} 26} & {\color[HTML]{212529} Zone 19 thermostat setpoint (°C).}     \\

{\color[HTML]{212529} Z20\_T\_Thermostat\_sp} & {\color[HTML]{212529} scalar}   & {\color[HTML]{212529} 16} & {\color[HTML]{212529} 26} & {\color[HTML]{212529} Zone 20 thermostat setpoint (°C).}     \\
 
{\color[HTML]{212529} Z25\_T\_Thermostat\_sp} & {\color[HTML]{212529} scalar}   & {\color[HTML]{212529} 16} & {\color[HTML]{212529} 26} & {\color[HTML]{212529} Zone 25 thermostat setpoint (°C).}     \\ 
						\end{longtable}
						
				\begin{longtable}[c]{ c  c  c  c  c }
			\caption{\label{table: seminar state space} State-space \(\mathcal{S}_{seminar}\) of the \textbf{\textit{Seminar-Thermostat} environment.}} \\
			
			\textbf{Variable}     & \textbf{Type} & \textbf{Lower Bound} & \textbf{Upper Bound} & \textbf{Description}\\
			\midrule
			\endfirsthead
			
			\multicolumn{5}{l}{{\tablename\  \thetable{} -- continued from previous page}} \\
			\textbf{Variable}     & \textbf{Type} & \textbf{Lower Bound} & \textbf{Upper Bound} & \textbf{Description}\\
			\midrule
			\endhead
			
			\hline
			\multicolumn{5}{r}{\textit{continued on next page}}
			\endfoot
			
			\hline \hline
			\endlastfoot
			{\color[HTML]{212529} Ext\_Irr}                    & {\color[HTML]{212529} scalar} & {\color[HTML]{212529} 0}     & {\color[HTML]{212529} 1000}   & {\color[HTML]{212529} Direct normal radiation (W/m2).}      \\
            {\color[HTML]{212529} Ext\_RH}                     & {\color[HTML]{212529} scalar} & {\color[HTML]{212529} 0}     & {\color[HTML]{212529} 100}    & {\color[HTML]{212529} Outdoor relative humidity (\%RH).}    \\
            {\color[HTML]{212529} Ext\_T}                      & {\color[HTML]{212529} scalar} & {\color[HTML]{212529} -15}   & {\color[HTML]{212529} 40}     & {\color[HTML]{212529} Outdoor temperature (°C).}            \\
            {\color[HTML]{212529} Ext\_P}                      & {\color[HTML]{212529} scalar} & {\color[HTML]{212529} 80000} & {\color[HTML]{212529} 130000} & {\color[HTML]{212529} Outdoor air pressure (Pa).}           \\
            {\color[HTML]{212529} Z01\_RH}                     & {\color[HTML]{212529} scalar} & {\color[HTML]{212529} 0}     & {\color[HTML]{212529} 100}    & {\color[HTML]{212529} Zone 1 relative humidity (\%RH).}     \\
            {\color[HTML]{212529} Z01\_T}                      & {\color[HTML]{212529} scalar} & {\color[HTML]{212529} 10}    & {\color[HTML]{212529} 40}     & {\color[HTML]{212529} Zone 1 temperature (°C).}             \\
            {\color[HTML]{212529} Z01\_T\_Thermostat\_sp\_out} & {\color[HTML]{212529} scalar} & {\color[HTML]{212529} 16}    & {\color[HTML]{212529} 26}     & {\color[HTML]{212529} Zone 1 thermostat setpoint (°C).}     \\
            {\color[HTML]{212529} Z02\_RH}                     & {\color[HTML]{212529} scalar} & {\color[HTML]{212529} 0}     & {\color[HTML]{212529} 100}    & {\color[HTML]{212529} Zone 2 relative humidity (\%RH).}     \\
            {\color[HTML]{212529} Z02\_T}                      & {\color[HTML]{212529} scalar} & {\color[HTML]{212529} 10}    & {\color[HTML]{212529} 40}     & {\color[HTML]{212529} Zone 2 temperature (°C).}             \\
            {\color[HTML]{212529} Z02\_T\_Thermostat\_sp\_out} & {\color[HTML]{212529} scalar} & {\color[HTML]{212529} 16}    & {\color[HTML]{212529} 26}     & {\color[HTML]{212529} Zone 2 thermostat setpoint (°C).}     \\
            {\color[HTML]{212529} Z03\_RH}                     & {\color[HTML]{212529} scalar} & {\color[HTML]{212529} 0}     & {\color[HTML]{212529} 100}    & {\color[HTML]{212529} Zone 3 relative humidity (\%RH).}     \\
            {\color[HTML]{212529} Z03\_T}                      & {\color[HTML]{212529} scalar} & {\color[HTML]{212529} 10}    & {\color[HTML]{212529} 40}     & {\color[HTML]{212529} Zone 3 temperature (°C).}             \\
            {\color[HTML]{212529} Z03\_T\_Thermostat\_sp\_out} & {\color[HTML]{212529} scalar} & {\color[HTML]{212529} 16}    & {\color[HTML]{212529} 26}     & {\color[HTML]{212529} Zone 3 thermostat setpoint (°C).}     \\
            {\color[HTML]{212529} Z04\_RH}                     & {\color[HTML]{212529} scalar} & {\color[HTML]{212529} 0}     & {\color[HTML]{212529} 100}    & {\color[HTML]{212529} Zone 4 relative humidity (\%RH).}     \\
            {\color[HTML]{212529} Z04\_T}                      & {\color[HTML]{212529} scalar} & {\color[HTML]{212529} 10}    & {\color[HTML]{212529} 40}     & {\color[HTML]{212529} Zone 4 temperature (°C).}             \\
            {\color[HTML]{212529} Z04\_T\_Thermostat\_sp\_out} & {\color[HTML]{212529} scalar} & {\color[HTML]{212529} 16}    & {\color[HTML]{212529} 26}     & {\color[HTML]{212529} Zone 4 thermostat setpoint (°C).}     \\
            {\color[HTML]{212529} Z05\_RH}                     & {\color[HTML]{212529} scalar} & {\color[HTML]{212529} 0}     & {\color[HTML]{212529} 100}    & {\color[HTML]{212529} Zone 5 relative humidity (\%RH).}     \\
            {\color[HTML]{212529} Z05\_T}                      & {\color[HTML]{212529} scalar} & {\color[HTML]{212529} 10}    & {\color[HTML]{212529} 40}     & {\color[HTML]{212529} Zone 5 temperature (°C).}             \\
            {\color[HTML]{212529} Z05\_T\_Thermostat\_sp\_out} & {\color[HTML]{212529} scalar} & {\color[HTML]{212529} 16}    & {\color[HTML]{212529} 26}     & {\color[HTML]{212529} Zone 5 thermostat setpoint (°C).}     \\
            {\color[HTML]{212529} Z06\_RH}                     & {\color[HTML]{212529} scalar} & {\color[HTML]{212529} 0}     & {\color[HTML]{212529} 100}    & {\color[HTML]{212529} Zone 6 relative humidity (\%RH).}     \\
            {\color[HTML]{212529} Z06\_T}                      & {\color[HTML]{212529} scalar} & {\color[HTML]{212529} 10}    & {\color[HTML]{212529} 40}     & {\color[HTML]{212529} Zone 6 temperature (°C).}             \\
            {\color[HTML]{212529} Z06\_T\_Thermostat\_sp\_out} & {\color[HTML]{212529} scalar} & {\color[HTML]{212529} 16}    & {\color[HTML]{212529} 26}     & {\color[HTML]{212529} Zone 6 thermostat setpoint (°C).}     \\
            {\color[HTML]{212529} Z08\_RH}                     & {\color[HTML]{212529} scalar} & {\color[HTML]{212529} 0}     & {\color[HTML]{212529} 100}    & {\color[HTML]{212529} Zone 8 relative humidity (\%RH).}     \\
            {\color[HTML]{212529} Z08\_T}                      & {\color[HTML]{212529} scalar} & {\color[HTML]{212529} 10}    & {\color[HTML]{212529} 40}     & {\color[HTML]{212529} Zone 8 temperature (°C).}             \\
            {\color[HTML]{212529} Z08\_T\_Thermostat\_sp\_out} & {\color[HTML]{212529} scalar} & {\color[HTML]{212529} 16}    & {\color[HTML]{212529} 26}     & {\color[HTML]{212529} Zone 8 thermostat setpoint (°C).}     \\
            {\color[HTML]{212529} Z09\_RH}                     & {\color[HTML]{212529} scalar} & {\color[HTML]{212529} 0}     & {\color[HTML]{212529} 100}    & {\color[HTML]{212529} Zone 9 relative humidity (\%RH).}     \\
            {\color[HTML]{212529} Z09\_T}                      & {\color[HTML]{212529} scalar} & {\color[HTML]{212529} 10}    & {\color[HTML]{212529} 40}     & {\color[HTML]{212529} Zone 9 temperature (°C).}             \\
            {\color[HTML]{212529} Z09\_T\_Thermostat\_sp\_out} & {\color[HTML]{212529} scalar} & {\color[HTML]{212529} 16}    & {\color[HTML]{212529} 26}     & {\color[HTML]{212529} Zone 9 thermostat setpoint (°C).}     \\
            {\color[HTML]{212529} Z10\_RH}                     & {\color[HTML]{212529} scalar} & {\color[HTML]{212529} 0}     & {\color[HTML]{212529} 100}    & {\color[HTML]{212529} Zone 10 relative humidity (\%RH).}    \\
            {\color[HTML]{212529} Z10\_T}                      & {\color[HTML]{212529} scalar} & {\color[HTML]{212529} 10}    & {\color[HTML]{212529} 40}     & {\color[HTML]{212529} Zone 10 temperature (°C).}            \\
            {\color[HTML]{212529} Z10\_T\_Thermostat\_sp\_out} & {\color[HTML]{212529} scalar} & {\color[HTML]{212529} 16}    & {\color[HTML]{212529} 26}     & {\color[HTML]{212529} Zone 10 thermostat setpoint (°C).}    \\
            {\color[HTML]{212529} Z11\_RH}                     & {\color[HTML]{212529} scalar} & {\color[HTML]{212529} 0}     & {\color[HTML]{212529} 100}    & {\color[HTML]{212529} Zone 11 relative humidity (\%RH).}    \\
            {\color[HTML]{212529} Z11\_T}                      & {\color[HTML]{212529} scalar} & {\color[HTML]{212529} 10}    & {\color[HTML]{212529} 40}     & {\color[HTML]{212529} Zone 11 temperature (°C).}            \\
            {\color[HTML]{212529} Z11\_T\_Thermostat\_sp\_out} & {\color[HTML]{212529} scalar} & {\color[HTML]{212529} 16}    & {\color[HTML]{212529} 26}     & {\color[HTML]{212529} Zone 11 thermostat setpoint (°C).}    \\
            {\color[HTML]{212529} Z13\_RH}                     & {\color[HTML]{212529} scalar} & {\color[HTML]{212529} 0}     & {\color[HTML]{212529} 100}    & {\color[HTML]{212529} Zone 12 relative humidity (\%RH).}    \\
            {\color[HTML]{212529} Z13\_T}                      & {\color[HTML]{212529} scalar} & {\color[HTML]{212529} 10}    & {\color[HTML]{212529} 40}     & {\color[HTML]{212529} Zone 13 temperature (°C).}            \\
            {\color[HTML]{212529} Z13\_T\_Thermostat\_sp\_out} & {\color[HTML]{212529} scalar} & {\color[HTML]{212529} 16}    & {\color[HTML]{212529} 26}     & {\color[HTML]{212529} Zone 13 thermostat setpoint (°C).}    \\
            {\color[HTML]{212529} Z14\_RH}                     & {\color[HTML]{212529} scalar} & {\color[HTML]{212529} 0}     & {\color[HTML]{212529} 100}    & {\color[HTML]{212529} Zone 14 relative humidity (\%RH).}    \\
            {\color[HTML]{212529} Z14\_T}                      & {\color[HTML]{212529} scalar} & {\color[HTML]{212529} 10}    & {\color[HTML]{212529} 40}     & {\color[HTML]{212529} Zone 14 temperature (°C).}            \\
            {\color[HTML]{212529} Z14\_T\_Thermostat\_sp\_out} & {\color[HTML]{212529} scalar} & {\color[HTML]{212529} 16}    & {\color[HTML]{212529} 26}     & {\color[HTML]{212529} Zone 14 thermostat setpoint (°C).}    \\
            {\color[HTML]{212529} Z15\_RH}                     & {\color[HTML]{212529} scalar} & {\color[HTML]{212529} 0}     & {\color[HTML]{212529} 100}    & {\color[HTML]{212529} Zone 15 relative humidity (\%RH).}    \\
            {\color[HTML]{212529} Z15\_T}                      & {\color[HTML]{212529} scalar} & {\color[HTML]{212529} 10}    & {\color[HTML]{212529} 40}     & {\color[HTML]{212529} Zone 15 temperature (°C).}            \\
            {\color[HTML]{212529} Z15\_T\_Thermostat\_sp\_out} & {\color[HTML]{212529} scalar} & {\color[HTML]{212529} 16}    & {\color[HTML]{212529} 26}     & {\color[HTML]{212529} Zone 15 thermostat setpoint (°C).}    \\
            {\color[HTML]{212529} Z18\_RH}                     & {\color[HTML]{212529} scalar} & {\color[HTML]{212529} 0}     & {\color[HTML]{212529} 100}    & {\color[HTML]{212529} Zone 18 relative humidity (\%RH).}    \\
            {\color[HTML]{212529} Z18\_T}                      & {\color[HTML]{212529} scalar} & {\color[HTML]{212529} 10}    & {\color[HTML]{212529} 40}     & {\color[HTML]{212529} Zone 18 temperature (°C).}            \\
            {\color[HTML]{212529} Z18\_T\_Thermostat\_sp\_out} & {\color[HTML]{212529} scalar} & {\color[HTML]{212529} 16}    & {\color[HTML]{212529} 26}     & {\color[HTML]{212529} Zone 18 thermostat setpoint (°C).}    \\
            {\color[HTML]{212529} Z19\_RH}                     & {\color[HTML]{212529} scalar} & {\color[HTML]{212529} 0}     & {\color[HTML]{212529} 100}    & {\color[HTML]{212529} Zone 19 relative humidity (\%RH).}    \\
            {\color[HTML]{212529} Z19\_T}                      & {\color[HTML]{212529} scalar} & {\color[HTML]{212529} 10}    & {\color[HTML]{212529} 40}     & {\color[HTML]{212529} Zone 19 temperature (°C).}            \\
            {\color[HTML]{212529} Z19\_T\_Thermostat\_sp\_out} & {\color[HTML]{212529} scalar} & {\color[HTML]{212529} 16}    & {\color[HTML]{212529} 26}     & {\color[HTML]{212529} Zone 19 thermostat setpoint (°C).}    \\
            {\color[HTML]{212529} Z20\_RH}                     & {\color[HTML]{212529} scalar} & {\color[HTML]{212529} 0}     & {\color[HTML]{212529} 100}    & {\color[HTML]{212529} Zone 20 relative humidity (\%RH).}    \\
            {\color[HTML]{212529} Z20\_T}                      & {\color[HTML]{212529} scalar} & {\color[HTML]{212529} 10}    & {\color[HTML]{212529} 40}     & {\color[HTML]{212529} Zone 20 temperature (°C).}            \\
            {\color[HTML]{212529} Z20\_T\_Thermostat\_sp\_out} & {\color[HTML]{212529} scalar} & {\color[HTML]{212529} 16}    & {\color[HTML]{212529} 26}     & {\color[HTML]{212529} Zone 20 thermostat setpoint (°C).}    \\
            {\color[HTML]{212529} Z21\_RH}                     & {\color[HTML]{212529} scalar} & {\color[HTML]{212529} 0}     & {\color[HTML]{212529} 100}    & {\color[HTML]{212529} Zone 21 relative humidity (\%RH).}    \\
            {\color[HTML]{212529} Z21\_T}                      & {\color[HTML]{212529} scalar} & {\color[HTML]{212529} 10}    & {\color[HTML]{212529} 40}     & {\color[HTML]{212529} Zone 21 temperature (°C).}            \\
            {\color[HTML]{212529} Z21\_T\_Thermostat\_sp\_out} & {\color[HTML]{212529} scalar} & {\color[HTML]{212529} 16}    & {\color[HTML]{212529} 26}     & {\color[HTML]{212529} Zone 21 thermostat setpoint (°C).}    \\
            {\color[HTML]{212529} Z22\_RH}                     & {\color[HTML]{212529} scalar} & {\color[HTML]{212529} 0}     & {\color[HTML]{212529} 100}    & {\color[HTML]{212529} Zone 22 relative humidity (\%RH).}    \\
            {\color[HTML]{212529} Z22\_T}                      & {\color[HTML]{212529} scalar} & {\color[HTML]{212529} 10}    & {\color[HTML]{212529} 40}     & {\color[HTML]{212529} Zone 22 temperature (°C).}            \\
            {\color[HTML]{212529} Z22\_T\_Thermostat\_sp\_out} & {\color[HTML]{212529} scalar} & {\color[HTML]{212529} 16}    & {\color[HTML]{212529} 26}     & {\color[HTML]{212529} Zone 22 thermostat setpoint (°C).}    \\
            {\color[HTML]{212529} Grid\_CO2}                   & {\color[HTML]{212529} scalar} & {\color[HTML]{212529} 0}     & {\color[HTML]{212529} 1000}     & {\color[HTML]{212529} Grid CO2 emission intensity (g/kWh).}
						\end{longtable}
						
			\begin{longtable}[c]{ c  c  c  c  c }
			\caption{\label{table: seminar action space} Action-space \(\mathcal{A}_{seminar}\) of the \textbf{\textit{Seminar-Thermostat} environment.}} \\
			
			\textbf{Action}     & \textbf{Type} & \textbf{Lower Bound} & \textbf{Upper Bound} & \textbf{Description}\\
			\midrule
			\endfirsthead
			
			\multicolumn{5}{l}{{\tablename\  \thetable{} -- continued from previous page}} \\
			
			\textbf{Action}     & \textbf{Type} & \textbf{Lower Bound} & \textbf{Upper Bound} & \textbf{Description}\\
			\midrule
			\endhead
			
			\hline
			\multicolumn{5}{r}{\textit{continued on next page}}
			\endfoot
			
			\hline \hline
			\endlastfoot
			
			{\color[HTML]{212529} Z01\_T\_Thermostat\_sp} & {\color[HTML]{212529} scalar} & {\color[HTML]{212529} 16} & {\color[HTML]{212529} 26} & {\color[HTML]{212529} Zone 1 thermostat setpoint (°C).}  \\
            {\color[HTML]{212529} Z02\_T\_Thermostat\_sp} & {\color[HTML]{212529} scalar} & {\color[HTML]{212529} 16} & {\color[HTML]{212529} 26} & {\color[HTML]{212529} Zone 2 thermostat setpoint (°C).}  \\
            {\color[HTML]{212529} Z03\_T\_Thermostat\_sp} & {\color[HTML]{212529} scalar} & {\color[HTML]{212529} 16} & {\color[HTML]{212529} 26} & {\color[HTML]{212529} Zone 3 thermostat setpoint (°C).}  \\
            {\color[HTML]{212529} Z04\_T\_Thermostat\_sp} & {\color[HTML]{212529} scalar} & {\color[HTML]{212529} 16} & {\color[HTML]{212529} 26} & {\color[HTML]{212529} Zone 4 thermostat setpoint (°C).}  \\
            {\color[HTML]{212529} Z05\_T\_Thermostat\_sp} & {\color[HTML]{212529} scalar} & {\color[HTML]{212529} 16} & {\color[HTML]{212529} 26} & {\color[HTML]{212529} Zone 5 thermostat setpoint (°C).}  \\
            {\color[HTML]{212529} Z06\_T\_Thermostat\_sp} & {\color[HTML]{212529} scalar} & {\color[HTML]{212529} 16} & {\color[HTML]{212529} 26} & {\color[HTML]{212529} Zone 6 thermostat setpoint (°C).}  \\
            {\color[HTML]{212529} Z08\_T\_Thermostat\_sp} & {\color[HTML]{212529} scalar} & {\color[HTML]{212529} 16} & {\color[HTML]{212529} 26} & {\color[HTML]{212529} Zone 8 thermostat setpoint (°C).}  \\
            {\color[HTML]{212529} Z09\_T\_Thermostat\_sp} & {\color[HTML]{212529} scalar} & {\color[HTML]{212529} 16} & {\color[HTML]{212529} 26} & {\color[HTML]{212529} Zone 9 thermostat setpoint (°C).}  \\
            {\color[HTML]{212529} Z10\_T\_Thermostat\_sp} & {\color[HTML]{212529} scalar} & {\color[HTML]{212529} 16} & {\color[HTML]{212529} 26} & {\color[HTML]{212529} Zone 10 thermostat setpoint (°C).} \\
            {\color[HTML]{212529} Z11\_T\_Thermostat\_sp} & {\color[HTML]{212529} scalar} & {\color[HTML]{212529} 16} & {\color[HTML]{212529} 26} & {\color[HTML]{212529} Zone 11 thermostat setpoint (°C).} \\
            {\color[HTML]{212529} Z13\_T\_Thermostat\_sp} & {\color[HTML]{212529} scalar} & {\color[HTML]{212529} 16} & {\color[HTML]{212529} 26} & {\color[HTML]{212529} Zone 13 thermostat setpoint (°C).} \\
            {\color[HTML]{212529} Z14\_T\_Thermostat\_sp} & {\color[HTML]{212529} scalar} & {\color[HTML]{212529} 16} & {\color[HTML]{212529} 26} & {\color[HTML]{212529} Zone 14 thermostat setpoint (°C).} \\
            {\color[HTML]{212529} Z15\_T\_Thermostat\_sp} & {\color[HTML]{212529} scalar} & {\color[HTML]{212529} 16} & {\color[HTML]{212529} 26} & {\color[HTML]{212529} Zone 15 thermostat setpoint (°C).} \\
            {\color[HTML]{212529} Z18\_T\_Thermostat\_sp} & {\color[HTML]{212529} scalar} & {\color[HTML]{212529} 16} & {\color[HTML]{212529} 26} & {\color[HTML]{212529} Zone 18 thermostat setpoint (°C).} \\
            {\color[HTML]{212529} Z19\_T\_Thermostat\_sp} & {\color[HTML]{212529} scalar} & {\color[HTML]{212529} 16} & {\color[HTML]{212529} 26} & {\color[HTML]{212529} Zone 19 thermostat setpoint (°C).} \\
            {\color[HTML]{212529} Z20\_T\_Thermostat\_sp} & {\color[HTML]{212529} scalar} & {\color[HTML]{212529} 16} & {\color[HTML]{212529} 26} & {\color[HTML]{212529} Zone 20 thermostat setpoint (°C).} \\
            {\color[HTML]{212529} Z21\_T\_Thermostat\_sp} & {\color[HTML]{212529} scalar} & {\color[HTML]{212529} 16} & {\color[HTML]{212529} 26} & {\color[HTML]{212529} Zone 21 thermostat setpoint (°C).} \\
            {\color[HTML]{212529} Z22\_T\_Thermostat\_sp} & {\color[HTML]{212529} scalar} & {\color[HTML]{212529} 16} & {\color[HTML]{212529} 26} & {\color[HTML]{212529} Zone 22 thermostat setpoint (°C).}
						\end{longtable}

\subsection{Grid Carbon Intensity} \label{appendix subsection: grid}
The \textit{Mixed-Use} and \textit{Office} environments do not provide native support for optimisation of grid carbon intensity $C[t]$. As such, we sourced grid carbon intensity data  from Electricity Maps, an open-source collection of global electrical grid data \cite{electricitymaps22}. Importantly, the data we sourced is for the same locale and year (Greece, 2017)  as the \textit{Energym} weather files to ensure any weather-grid dependencies are preserved. We append $C[t]$ to the \textit{Energym} state spaces.  

\section{Agents}
\subsection{Rule Based Controller (RBC)} \label{appendix: rbc}
The RBC's objective is to maintain temperature at 22\(^\circ\)C in the building. To achieve this it selects actions as follows:
\begin{equation} \label{eq: rbc policy}
	a_t^i[t+1]= 
	\begin{cases}
		a_t^i[t] + 0.5^\circ C:& T_{obs}^i \leq 21.2^\circ C \\
		a_t^i[t] - 0.5^\circ C:& T_{obs}^i \geq 22.8^\circ C \\
		a_t^i[t]:& 21.2^\circ C \leq T_{obs}^i \leq 22.8^\circ C 
	\end{cases}
\end{equation}

\subsection{Random Walk Exploration} \label{appendix: lazic}
During system identification, Lazic et. al (2018)'s agent performs a uniform random walk in each control variable bounded to a safe range over the action space informed by historical data if available, or initialised conservatively and expanded if no data is available. Stepwise changes in action selection are bounded to mitigate potential hardware damage that may be caused by large swings in chosen actions.

Formally their policy as follows. Let $a^i[t]$ be the value of action $i$ at timestep $t$, with $i \in \mathcal{A}$. Let [$a_{min}^i$, $a_{max}^i$] be the safe operating range for action $i$, and let $\Delta^i$ be the maximum allowable change in action $a^i$ between timesteps. The system identification strategy, $\pi_{system ID}$, is therefore:
\begin{equation} \label{eq: system ID}
a^i[t+1] = \max\left(a_{min}^i, \min(a_{max}^i, a^i[t] + v^i)\right), \; v^i \sim \text{Uniform}\left(- \Delta ^i, \Delta^i\right) \; .
\end{equation}

Model parameters are updated iteratively when the number of timesteps observed in the environment equals the batch size of the model. 

\subsection{Hyperparameters} \label{appendix: hyperparameters}
\begin{longtable}[c]{ l  c  c }
			\caption{\label{table: model-based hyperparameters} \textbf{PEARL and MPC-DNN hyperparameters.}}\\

			\textbf{Hyperparameter}                              & \textbf{PEARL} & \textbf{MPC-DNN} \\
			\midrule
			\endfirsthead
			
			\multicolumn{3}{l}{{\tablename\  \thetable{} -- continued from previous page}} \\
			
			\textbf{Hyperparameter}                              & \textbf{PEARL} & \textbf{MPC-DNN} \\
			\midrule
			\endhead
			
			\hline
			\multicolumn{3}{r}{\textit{continued on next page}}
			\endfoot
			
			\hline \hline
			\endlastfoot
			
			Horizon minutes ($H$)                                          & 300    & 300        \\
			History ($h$)                                                   & 2 & 2 \\
		Adam stepsize ($\alpha$)                & 0.0003      & 0.0003   \\
		Num. epochs                                          & 25 & 25             \\
		Minibatch size                                       & 32  & 32           \\
		No. of candidate actions ($U$)                                         & 25 & 25          \\
		No. of particles ($P$)                                         & 10 & n/a          \\
		MPPI Iterations ($N$)                                   & 5 & -- \\
		Network Fully Connected Layers                 & 5   & 5           \\
		Network Layer Dimensions                       & 200  & 200          \\
		Network Activation Function                    & Tanh  & Tanh         \\
		Ensemble Models ($K$)                & 5      & n/a       \\
		System ID Minutes ($C$)                   & 180     & n/a      \\
		Reward Emissions Parameter ($\phi$)       & 0.001           & 0.001  \\
		Discount ($\gamma$)                                         & 1 & 1           \\
		\end{longtable}
		
\begin{longtable}[c]{ l  c }
			\caption{\label{table: ppo hyperparameters} \textbf{PPO hyperparameters.}}\\

			\textbf{Hyperparameter}                              & \textbf{PPO} \\
			\midrule
			\endfirsthead
			
			\multicolumn{2}{l}{{\tablename\  \thetable{} -- continued from previous page}} \\
			
			\textbf{Hyperparameter}     & \textbf{PPO}\\
			\midrule
			\endhead
			
			\hline
			\multicolumn{2}{r}{\textit{continued on next page}}
			\endfoot
			
			\hline \hline
			\endlastfoot
					Batch Size                                           & 32            \\
					History ($h$)                                                   & 2 \\
		Adam stepsize ($\alpha$)                & 0.0003         \\
		Epochs per update                                          & 50             \\
		Discount ($\gamma$)                                         & 0.99           \\
		GAE parameter ($\lambda$)                                    & 0.95           \\
		Clipping parameter									& 0.2            \\
		Actor Network Fully Connected Layers                 & 5              \\
		Actor Network Layer Dimensions                       & 200            \\
		Actor Network Activation Function                    & Tanh           \\
		Critic Network Fully Connected Layers                & 5           \\
		Critic Network Layer Dimensions                      & 200            \\
		Critic Network Activation Function                   & Tanh           \\
		Reward Emissions Parameter ($\phi$)       & 0.001              \\
		\end{longtable}

\begin{longtable}[c]{ l  c }
			\caption{\label{table: sac hyperparameters} \textbf{SAC hyperparameters.}}\\

			\textbf{Hyperparameter}                              & \textbf{SAC} \\
			\midrule
			\endfirsthead
			
			\multicolumn{2}{l}{{\tablename\  \thetable{} -- continued from previous page}} \\
			
			\textbf{Hyperparameter}     & \textbf{SAC}\\
			\midrule
			\endhead
			
			\hline
			\multicolumn{2}{r}{\textit{continued on next page}}
			\endfoot
			
			\hline \hline
			\endlastfoot
					Batch Size                                           & 256            \\
					History ($h$)                                                   & 2 \\
		Adam stepsize ($\alpha$)                & 0.0003         \\
		Epochs per update                                          & 20             \\
		Discount ($\gamma$)                                         & 0.99           \\
		Target Smoothing Coefficient ($\tau$)                                    & 0.005           \\
		Hidden Layers (all networks)                 & 2              \\
		Layer Dimensions                       & 256            \\
		Activation Function                    & ReLU           \\
		Reward Scale                            & 2             \\
		Reward Emissions Parameter ($\phi$)       & 1              \\
		Reward Temperature Parameter ($\theta$) & 1000    
		\end{longtable}
		
\begin{longtable}[c]{ l  c  c  c }
			\caption{\label{table: sac oracle hyperparameters} \textbf{SAC Oracle hyperparameters.}}\\

			\textbf{Hyperparameter}                              & \textbf{Mixed-use} & \textbf{Offices} & \textbf{Seminar Centre} \\
			\midrule
			\endfirsthead
			
			\multicolumn{2}{l}{{\tablename\  \thetable{} -- continued from previous page}} \\
			
			\textbf{Hyperparameter}                              & \textbf{Mixed-use} & \textbf{Offices} & \textbf{Seminar Centre} \\
			\midrule
			\endhead
			
			\hline
			\multicolumn{2}{r}{\textit{continued on next page}}
			\endfoot
			
			\hline \hline
			\endlastfoot
					Batch Size                                           & 256   & 256 & 1024        \\
					History ($h$)                                                   & 2 & 2 & 2 \\
		Adam stepsize ($\alpha$)                & 0.0003  & 0.0015 &   0.0080    \\
		Epochs per update                                          & 1    & 1 & 1         \\
		Discount ($\gamma$)                                         & 0.9864  & 0.7539 & 0.8533          \\
		Target Smoothing Coefficient ($\tau$)                                    & 0.07815    & 0.2532 & 0.2890       \\
		Hidden Layers (all networks)                 & 2 & 2 & 2              \\
		Layer Dimensions                       & 256  & 256 & 256          \\
		Activation Function                    & ReLU   & ReLU & ReLU        \\
		Reward Scale                            & 2 & 20  &    20              \\
		Reward Emissions Parameter ($\phi$)       & 1 & 1 & 1             \\
		Reward Temperature Parameter ($\theta$) & 1000   & 1000 & 1000
		\end{longtable}

\subsection{Reward Curves} \label{appendix: rewards}
\begin{figure}[ht]
    \centering
    \includegraphics[width=0.73\textwidth]{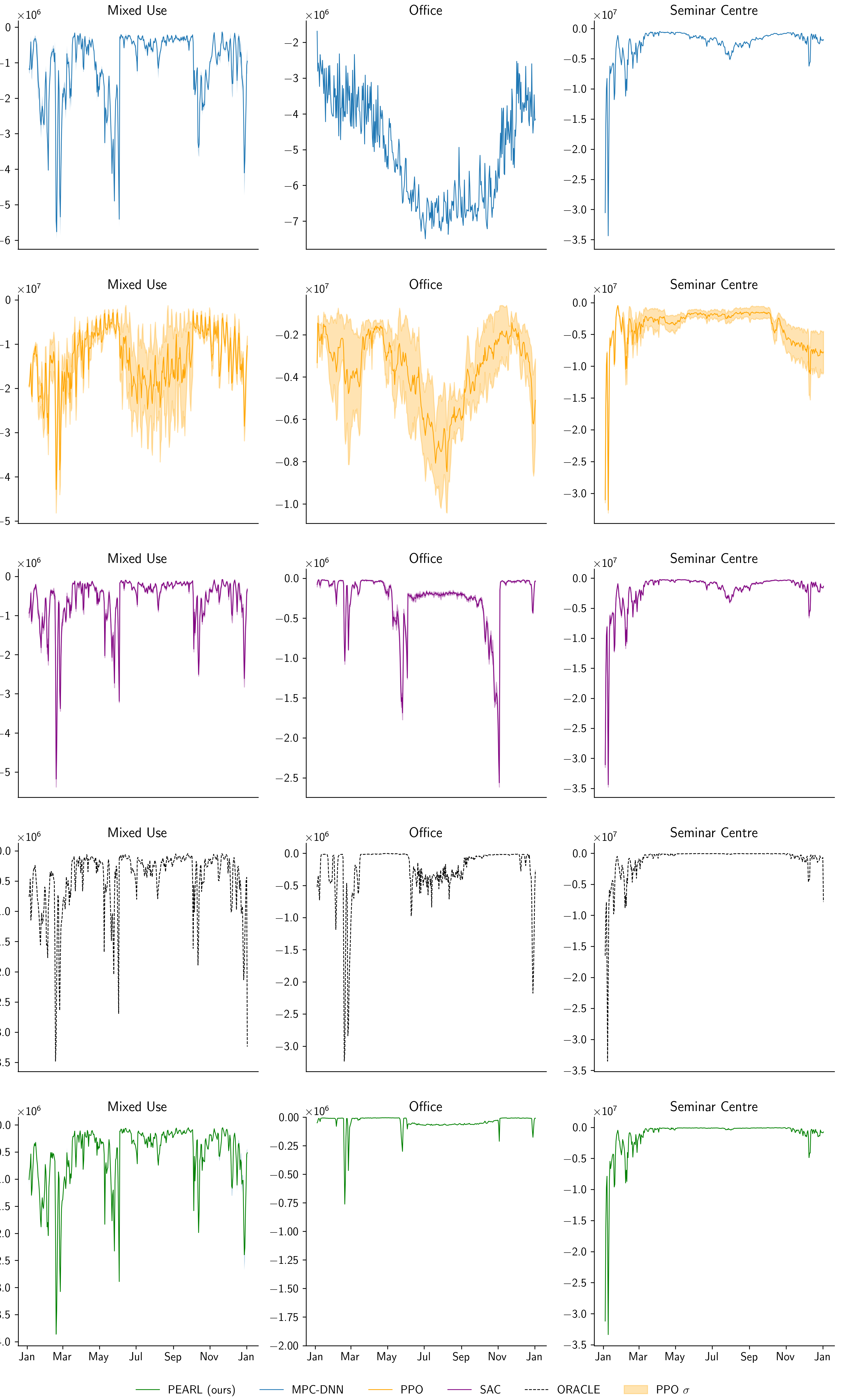}
    \caption{\textbf{Rewards.} Daily mean reward for all RL agents across each \textit{Energym} environment. Note the differing y-axes.}
    \label{fig: reward curves}
\end{figure}

Mean daily reward for 3 runs of each agent in each environment are illustrated in Figure \ref{fig: reward curves}. 

\section{Hardware}
All experiments were completed locally on a machine with the following specifications:
\begin{itemize}
    \item OS: Windows 10
    \item RAM: 32GB 4X8GB 2666MHz DDR4
    \item CPU: Intel Core™ i7-8700 (6 Cores/12MB/12T/up to 4.6GHz/65W)
    \item GPU: 1X NVIDIA GeForce RTX 3050 Ti 
\end{itemize}

\end{document}